\pdfoutput=1

\documentclass[11pt]{article}

\usepackage{acl}

\usepackage{times}
\usepackage{latexsym}
\usepackage{times}
\usepackage{latexsym}
\usepackage{CJKutf8}
\usepackage{amsmath}
\usepackage{verbatim}
\usepackage{booktabs}
\usepackage{amssymb}
\usepackage{pifont}
\newcommand{\cmark}{\ding{52}}%
\newcommand{\xmark}{\ding{56}}%
\usepackage{stfloats}
\usepackage{multicol}
\usepackage{float}
\usepackage{graphicx}
\usepackage{subfigure}
\usepackage{multirow}
\usepackage{array}
\newcommand{\PreserveBackslash}[1]{\let\temp=\\#1\let\\=\temp}

\newcolumntype{C}[1]{>{\PreserveBackslash\centering}p{#1}}
\newcolumntype{R}[1]{>{\PreserveBackslash\raggedleft}p{#1}}
\newcolumntype{L}[1]{>{\PreserveBackslash\raggedright}p{#1}}
\usepackage{diagbox}
\usepackage[noend]{algpseudocode}
\usepackage[ruled,linesnumbered,lined,commentsnumbered]{algorithm2e}
\usepackage{booktabs}
\usepackage{amsfonts}
\usepackage{bbm}
\usepackage{mathtools}
\usepackage{enumitem}

\usepackage[T1]{fontenc}

\usepackage[utf8]{inputenc}
\usepackage{microtype}

\usepackage{inconsolata}

\title{StreamSpeech: Simultaneous Speech-to-Speech Translation with Multi-task Learning}

\author{Shaolei Zhang$^{1,3}$, Qingkai Fang$^{1,3}$, Shoutao Guo$^{1,3}$, Zhengrui Ma$^{1,3}$,\\ \textbf{Min Zhang}$^{4}$\textbf{, Yang Feng$^{1,2,3}$}\thanks{Corresponding author: Yang Feng}\\
$^1$Key Laboratory of Intelligent Information Processing,\\ Institute of Computing Technology, Chinese Academy of Sciences (ICT/CAS) \\
$^2$Key Laboratory of AI Safety, Chinese Academy of Sciences \\
$^3$University of Chinese Academy of Sciences, Beijing, China\\
$^4$School of Future Science and Engineering, Soochow University\\
\texttt{\href{mailto:zhangshaolei20z@ict.ac.cn}{zhangshaolei20z@ict.ac.cn}},~\texttt{\href{mailto:zhangminmt@hotmail.com}{zhangminmt@hotmail.com}},~\texttt{\href{mailto:fengyang@ict.ac.cn}{fengyang@ict.ac.cn}}
}

\begin{document}
\maketitle
\begin{abstract}

Simultaneous speech-to-speech translation (Simul-S2ST, a.k.a streaming speech translation) outputs target speech while receiving streaming speech inputs, which is critical for real-time communication. 
Beyond accomplishing translation between speech, Simul-S2ST requires a policy to control the model to generate corresponding target speech at the opportune moment within speech inputs, thereby posing a double challenge of translation and policy. In this paper, we propose \emph{StreamSpeech}, a direct Simul-S2ST model that jointly learns translation and simultaneous policy in a unified framework of multi-task learning. 
Adhering to a multi-task learning approach, StreamSpeech can perform offline and simultaneous speech recognition, speech translation and speech synthesis via an ``All-in-One'' seamless model. Experiments on CVSS benchmark demonstrate that StreamSpeech achieves state-of-the-art performance in both offline S2ST and Simul-S2ST tasks. Besides, StreamSpeech is able to present high-quality intermediate results (i.e., ASR or translation results) during simultaneous translation process, offering a more comprehensive real-time communication experience\footnote{Code: \url{https://github.com/ictnlp/StreamSpeech}\\Project Page: \hypersetup{urlcolor=magenta}\href{https://ictnlp.github.io/StreamSpeech-site/}{https://ictnlp.github.io/StreamSpeech-site/}}.

\end{abstract}

\section{Introduction}
Simultaneous speech-to-speech translation (Simul-S2ST), which involves generating target speech while concurrently receiving streaming speech inputs \citep{iwslt-2023-international,gpt-4o}, has become an indispensable technology for low-latency communication in various scenarios, such as international conferences, live broadcasts and online subtitles. To produce high-quality translated speech under low latency, Simul-S2ST requires a policy to determine the optimal moments to start translating within the streaming speech inputs (i.e., READ action) and subsequently generate coherent target speech outputs (i.e., WRITE action) \citep{gu-etal-2017-learning}.

Existing simultaneous translation methods focus on text-to-text (Simul-T2TT) \citep{ma-etal-2019-stacl,Arivazhagan2019,zhang2023hidden} and speech-to-text translation (Simul-S2TT) \citep{ren-etal-2020-simulspeech,chen-etal-2021-direct,zeng-etal-2021-realtrans,zhang-feng-2023-end}. Such methods typically require cascading external modules such as speech recognition (ASR) and text-to-speech synthesis (TTS) to accomplish Simul-S2ST. However, this cascaded approach tends to amplify inference errors progressively between modules \citep{zhang-etal-2022-learning,ma-etal-2020-simulmt}, and also impedes the joint optimization of various modules \citep{zhang2023unified}. To address these issues, developing a direct Simul-S2ST model is imperative, particularly given the promising potential exhibited by end-to-end models such as SeamlessM4T \citep{communication2023seamless} and GPT-4o \citep{gpt-4o}.

\begin{figure}[t]
    \centering
    \vspace{5mm}
    \includegraphics[width=0.499\textwidth]{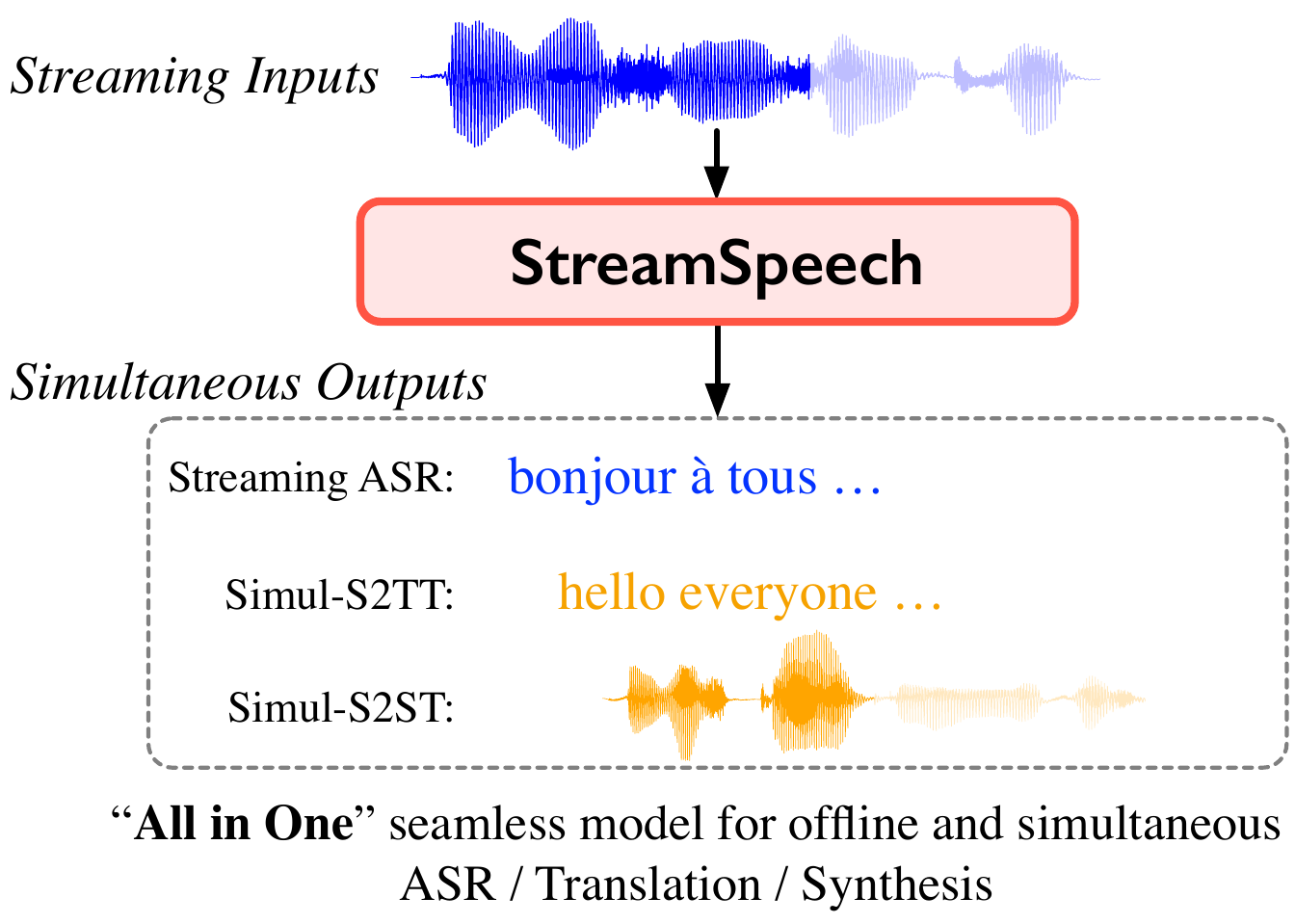}
    \caption{StreamSpeech is an ``All in One'' seamless model for multiple offline and simultaneous tasks.}
    \label{fig:ill}
\end{figure}

Direct speech-to-speech translation (S2ST) is already highly challenging, and the goal of accomplishing it simultaneously (Simul-S2ST) further exacerbates the difficulty. For translation, speech involves more diverse representation due to additional features such as timbre and intonation \citep{pmlr-v162-jia22b}, which renders directly translating source speech to target speech challenging. In simultaneous scenarios, beyond translation, the model additionally requires a policy to identify the appropriate translating moments, which is non-trivial to directly accomplish due to the continuous nature and uncertain duration of speech \citep{zhang-feng-2023-end}. Therefore, Simul-S2ST faces the double challenges of translation and policy.

To address the challenges of translation and policy, we aim to introduce textual information of both source and target speech to guide Simul-S2ST, which can not only provide intermediate supervision for translation but also guide the policy by establishing an alignment between source and target speech with text as a bridge. Specifically, a reasonable policy should control the model to wait until recognizing the presence of text in the received speech (READ), facilitated by the alignment between source speech and source text. Subsequently, the model should generate target speech corresponding to inputs (WRITE), which can be guided by the alignments from the source speech to target text and from the target text to target speech.

\begin{figure*}[t]
    \centering
    \includegraphics[width=0.996\textwidth]{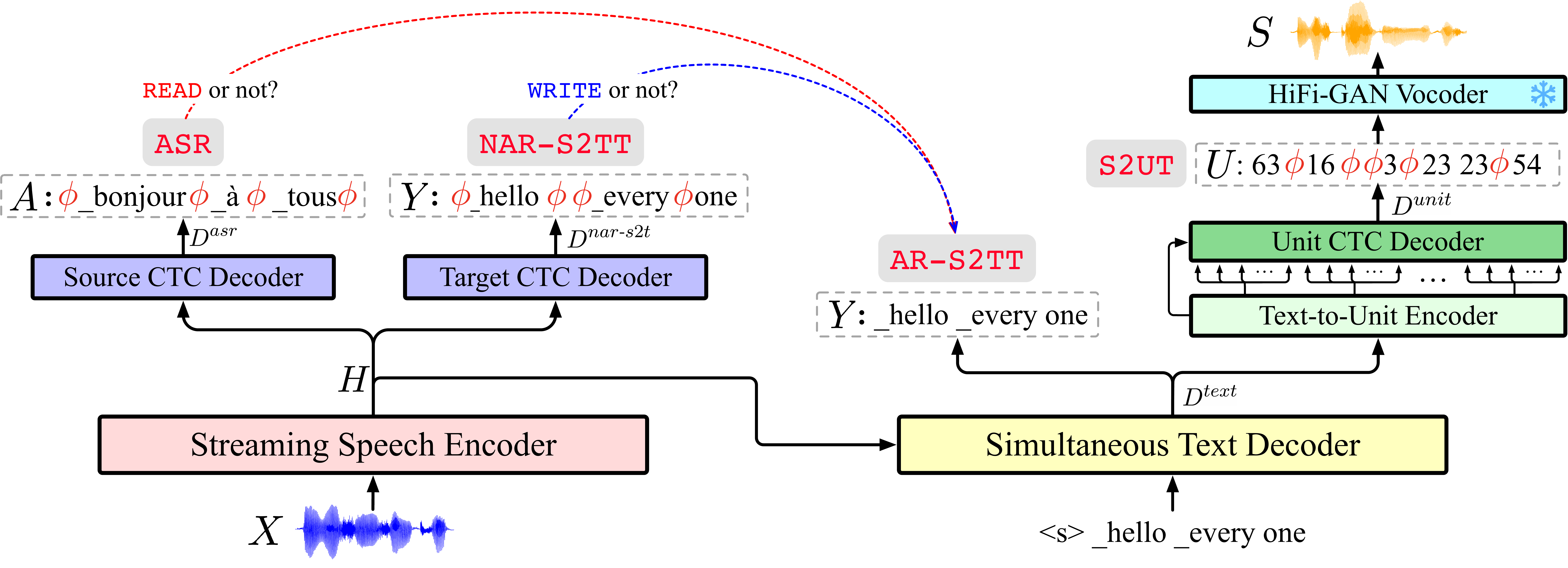}
    \caption{StreamSpeech employs two-pass architecture that first converts source speech into target text hidden states $D^{text}$ (autoregressive speech-to-text translation, AR-S2TT) and then generates target speech via non-autoregressive text-to-unit generation. The source/target/unit CTC decoders are introduced to learn alignments via multiple tasks of speech recognition (ASR), non-autoregressive speech-to-text translation (NAR-S2TT) and speech-to-unit translation (S2UT), accordingly guiding StreamSpeech when to start recognizing, translating and synthesizing.}
    \label{fig:streamspeech}
\end{figure*}

Given the pivotal role of text in both translation and alignment-guided policy, we propose \emph{StreamSpeech}, a direct Simul-S2ST model that jointly learns translation and policy in a unified framework of multi-task learning. StreamSpeech employs the advanced two-pass architecture \citep{inaguma-etal-2023-unity,pmlr-v162-jia22b}, which first translates source speech into target text hidden states, and then converts the text hidden states into target speech. Furthermore, we introduce multiple connectionist temporal classification (CTC) \citep{ctc} decoders and optimize them via auxiliary tasks of ASR and S2TT, thereby providing intermediate supervision for translation and meanwhile learning alignments to guide policy. All modules in StreamSpeech are jointly optimized through multi-task learning, facilitating jointly learning of translation and policy. Experiments show that StreamSpeech exhibits adaptability to different latency and achieves state-of-the-art performance on both offline S2ST and Simul-S2ST tasks.

\section{Background}
\label{sec:back}
\textbf{Speech-to-Speech Translation (S2ST)}\quad The corpus we used for speech-to-speech translation (S2ST) task is denoted as quadruple: $\mathcal{D}\!=\!\{\left(X,A,Y,S\right)\}$, where $X\!=\!\left(x_{1},\cdots,x_{ \left|X \right|}\right)$ is the source speech, $A\!=\!\left(a_{1},\cdots,a_{ \left|A \right|}\right)$ is the transcribed text of source speech, $Y\!=\!\left(y_{1},\cdots,y_{ \left|Y \right|}\right)$ is the target text, $S\!=\!\left(s_{1},\cdots,s_{ \left|S \right|}\right)$ is the target speech. The current mainstream methods for S2ST \citep{inaguma-etal-2023-unity} extract a discrete unit sequence $U\!=\!\left(u_{1},\cdots,u_{ \left|U \right|}\right)$ from the target speech, and employ a two-pass architecture, where both the first and second passes use autoregressive encoder-decoder. The first pass transforms the source speech to target text hidden states, and the second pass generates the discrete unit sequence based on the text hidden states, followed by a pre-trained unit-based HiFi-GAN vocoder \citep{NEURIPS2020_c5d73680} for target speech synthesis. In addition to the primary speech-to-unit translation (S2UT, $X\rightarrow U$), an auxiliary speech-to-text translation task (S2TT, $X\rightarrow Y$) is introduced to provide supervision.

\textbf{Connectionist Temporal Classification (CTC)}\quad \citep{ctc} CTC is a technique used to model alignment between two sequences of unequal lengths. For a longer input sequence $\mathcal{X}$, CTC decoder generates a same-length sequence $\mathcal{Z}$ containing repeated and blank tokens $\phi$, which is subsequently shortened by merging consecutively repeated tokens and removing blank tokens $\phi$ via collapsing function $\Pi\left ( \cdot \right )$. During training, given the ground-truth sequence $\mathcal{Y}$, CTC loss is calculated on all sequences $\mathcal{Z}$ that can be reduced to $\mathcal{Y}$ via the collapsing function:
\begin{gather}
    \mathrm{CTC}(\mathcal{X},\mathcal{Y})=-\log \!\! \sum_{\mathcal{Z}\in \Pi^{-1} \left (\mathcal{Y}  \right ) }\!\!p\left ( \mathcal{Z} \mid \mathcal{X} \right ).
\end{gather}

\section{StreamSpeech}

\subsection{Architecture}

The overall architecture of StreamSpeech is illustrated in Figure \ref{fig:streamspeech}. StreamSpeech consists of three parts: streaming speech encoder, simultaneous text decoder and synchronized text-to-unit generation module. Multiple CTC decoders are introduced to learn the alignments through auxiliary tasks and accordingly guide the policy.

\textbf{Streaming Speech Encoder}\quad 
Conformer architecture exhibits remarkable advantages in speech modeling by stacking attention modules and convolutional modules \citep{gulati20_interspeech}, while it struggles to model the streaming speech inputs, primarily due to the bi-directional self-attention and convolutional operations involving the entire sequence's receptive field. To this end, we propose \emph{chunk-based Conformer}, aiming to endow the Conformer architecture with the capability to encode streaming inputs while retaining the bi-directional encoding within local chunk.

\begin{figure}[t]
    \centering
    \includegraphics[width=0.45\textwidth]{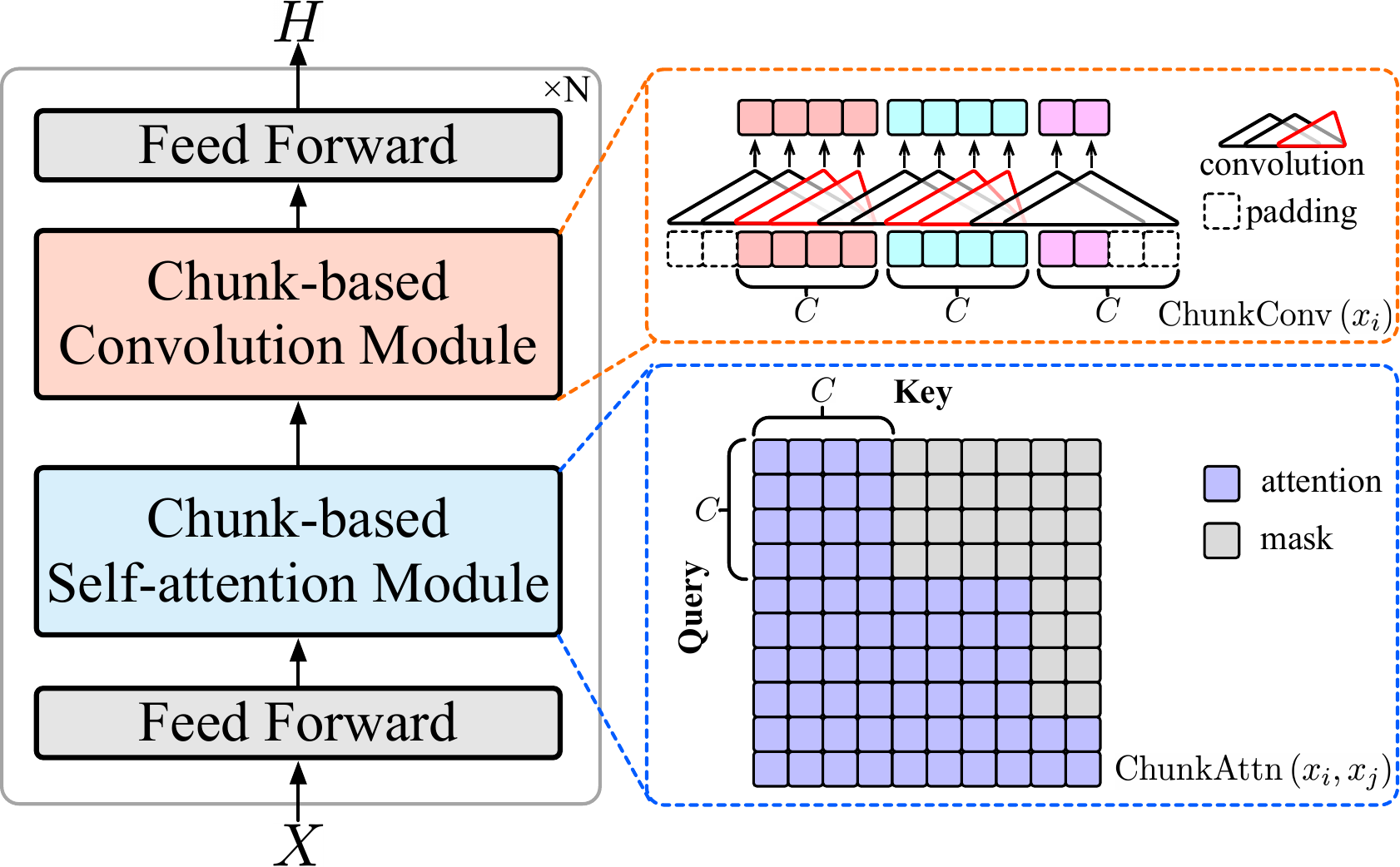}
    \caption{Architecture of chunk-based Conformer.}
    \label{fig:chunk_wise_conformer}
\end{figure}

Figure \ref{fig:chunk_wise_conformer} shows the architecture of chunk-based Conformer. First of all, the raw speech inputs are converted to speech features (we use filterbank features \citep{Povey:192584} in our work), where each speech feature typically corresponds to a 40$ms$ duration. Chunk-based Conformer divides the streaming speech into chunks, each containing $C$ speech features, where $C$ is a hyperparameter controlling the chunk size. In the chunk-based Conformer, self-attention and convolution operations are both bidirectional within a chunk and unidirectional between chunks, thereby handling the streaming inputs. For chunk-based self-attention, feature $x_{i}$ pays attention to the features $x_{j}$ that are located in the same and previous chunks, calculated as:
\begin{gather}
\begin{aligned}
     &\mathrm{ChunkAttn}\left(x_{i},x_{j}\right )\\
     &\;\;\;\;\;\;\;= \begin{cases}
\mathrm{Attn}\left (x_{i},x_{j}  \right ) & \text{ if }\; j\leq \left \lceil \frac{i}{C}\right \rceil\times C \\
0 & \text{ otherwise }\; 
\end{cases},
\end{aligned}
\end{gather}
where $\mathrm{Attn}\left (x_{i},x_{j}  \right )$ is standard multi-head attention \citep{NIPS2017_7181}, and $\left \lceil \cdot\right \rceil$ is ceiling operation. For chunk-based convolution, the convolution operation with kernel size $k$ is truncated at the upper bound of the chunk, calculated as:
\begin{align}
    &\mathrm{ChunkConv}\left ( x_{i} \right )=\\
    &\mathrm{Conv}\!\left ( x_{i-\frac{k-1}{2}},\!\cdots\!,x_{i},\!\cdots\!,x_{\min(i+\frac{k-1}{2},\left \lceil \!\frac{i}{C}\!\right \rceil\times C)} \right ). \nonumber
\end{align}
where $\left \lceil \frac{i}{C}\right \rceil\times C$ is the upper bound of the chunk that $x_{i}$ is located in. In implementation, chunk-based convolution can be computed in parallel through a mask operation (mask out those truncated positions). Through the streaming encoder, the source speech hidden states are calculated, denoted as $H=\left(h_{1},\cdots,h_{ \left|H \right|}\right)$. With chunk-based Conformer, the streaming speech encoder not only fulfills the need for streaming encoding but also conducts local bi-directional encoding of speech.

\textbf{Simultaneous Text Decoder}\quad After streaming encoder, text decoder simultaneously generates target text $Y$ by attending the source speech hidden states $H$. To achieve this, StreamSpeech requires a policy to decide when to generate each target token (i.e., how many speech states can the decoder attend to.). A reasonable policy should ensure that the model waits until recognizing the source text in the source speech (READ), and then generates the corresponding target text (WRITE).

To this end, we aim to align the source and target text to the speech inputs, thereby guiding ``READ or not'' and ``WRITE or not'' respectively. Considering the length difference between speech and text sequences, we align them via CTC decoder (refer to Sec.\ref{sec:back}). Specifically, we introduce a source CTC decoder $\texttt{CTCDec}^{\texttt{A}}(\cdot)$ and a target CTC decoder $\texttt{CTCDec}^{\texttt{Y}}(\cdot)$ at the top of the streaming speech encoder to generate source and target text:
\begin{align}
    D^{asr}=&\;\texttt{CTCDec}^{\texttt{A}}(H),\\ D^{nar\text{-}s2tt}=&\;\texttt{CTCDec}^{\texttt{Y}}(H), \label{eq:ctc_decoder}
\end{align}
and optimize them through the auxiliary tasks of speech recognition (ASR, $X\rightarrow A$) and non-autoregressive speech-to-text translation (NAR-S2TT, $X\rightarrow Y$), via CTC loss respectively:
\begin{align}
    \mathcal{L}_{asr}=&\;\mathrm{CTC}(D^{asr},A), \\ \mathcal{L}_{nar\text{-}s2tt}=&\;\mathrm{CTC}(D^{nar\text{-}s2tt},Y).
\end{align}

With CTC decoders, the source and target text are aligned to source speech. Accordingly, StreamSpeech starts translating upon the source CTC decoder recognizing a new source token from source speech, and then autoregressively generates target tokens that align to the received speech within target CTC decoder\footnote{NAR-S2TT can achieve well 1-gram token accuracy, but its translations are often less smooth compared to AR-S2TT. Therefore, StreamSpeech adopts NAR-S2TT to capture alignment and guide the policy, while still leveraging AR-S2TT to generate target tokens for better translation quality.}. Therefore, we calculate the number of source tokens and target tokens aligned to the current speech inputs $X_{\leq j}$, denoted as $\mathcal{N}^{asr}_{j}$ and $\mathcal{N}^{nar\text{-}s2t}_{j}$, respectively. Note that during training, we calculate the expected number of tokens contained in the CTC sequence, where the specific calculation is introduced in Appendix \ref{app:ctc}.

Given $\mathcal{N}^{asr}_{j}$ and $\mathcal{N}^{nar\text{-}s2tt}_{j}$, StreamSpeech autoregressively generates target token $y_{i}$ after receiving speech $X_{\leq g\left ( i \right )}$, where $g\left ( i \right )$ is defined as:
\begin{gather}
    g\left ( i \right )=\underset{\left\{ j\, \mid \,\mathcal{N}^{asr}_{j-1}<\mathcal{N}^{asr}_{j} \right\}}{\mathrm{argmin}}\left (\mathcal{N}^{nar\text{-}s2tt}_{j}\geq i  \right ).
\end{gather}
$\mathcal{N}^{asr}_{j-1}<\mathcal{N}^{asr}_{j} $ ensures that StreamSpeech starts translating when a new source token is recognized, and $(\mathcal{N}^{nar\text{-}s2tt}_{j}\geq i )$ ensures that StreamSpeech generates those target tokens that align to the received speech. Based on the policy guided by the alignments derived from ASR and NAR-S2TT, simultaneous text decoder generates $y_{i}$ after receiving speech $X_{\leq g(i)}$, and optimized via cross-entropy loss on autoregressive speech-to-text translation (AR-S2TT, $X\rightarrow Y$):
\begin{gather}
    \mathcal{L}_{ar\text{-}s2tt}\!=\!-\frac{1}{\left| Y\right|}\!\sum_{i=1}^{\left| Y\right|} \log p\left (y_{i} \mid X_{\leq g(i)},Y_{<i} \right ). \label{eq:s2t}
\end{gather}

\textbf{Non-autoregressive Text-to-Unit Generation}\quad To synchronously generate the corresponding unit for the currently generated text, StreamSpeech employs a non-autoregressive text-to-unit (T2U) architecture \citep{gu2018nonautoregressive}, comprising a T2U encoder and a unit CTC decoder. T2U encoder takes the hidden state $D^{text}$ from the simultaneous text decoder as inputs. For the unit CTC decoder, considering that unit sequences $U$ are often longer than text sequences $Y$, we upsample the T2U encoder outputs $r$ times as the decoder inputs, where the $i^{th}$ input corresponds to $D^{text}_{\left \lceil i/r\right \rceil}$. Then unit CTC decoder generates the unit sequence $U$ non-autoregressively by attending to those T2U encoder outputs located before $D^{text}_{\left \lceil i/r \right \rceil}$. Formally, the output $D^{unit}$ of unit CTC decoder $\texttt{CTCDec}^{\texttt{U}}$ is calculated as:
\begin{gather}
    D^{unit}_{i}=\texttt{CTCDec}^{\texttt{U}}\left (D^{text}_{\leq \left \lceil{i}/{r}\right \rceil}  \right). \label{eq:t2u}
\end{gather}
NAR T2U generation is optimized on speech-to-unit translation task (S2UT, $S\rightarrow U$) via CTC loss:
\begin{gather}
    \mathcal{L}_{s2ut}= \mathrm{CTC}(D^{unit},U).
\end{gather}

Finally, a unit-based HiFi-GAN vocoder \citep{NEURIPS2020_c5d73680} is used to synthesize target speech based on the generated units. Note that the HiFi-GAN vocoder is often pre-trained and frozen.

\subsection{Training}

All tasks involved in StreamSpeech are jointly optimized via multi-task learning in an end-to-end manner, and the total training objective $\mathcal{L}$ encompasses the losses of S2UT, AR-S2TT, ASR, and NAR-S2TT tasks:
\begin{gather}
\mathcal{L}=\mathcal{L}_{s2ut}+\mathcal{L}_{ar\text{-}s2tt}+\mathcal{L}_{asr}+\mathcal{L}_{nar\text{-}s2tt}.
\end{gather}
Multi-task learning effectively integrates the learning of simultaneous policy and translation into a unify framework. Besides, the high-quality intermediate results of auxiliary tasks, such as ASR and AR-S2TT, can also be displayed to users during inference as supplementary products.

\textbf{Multi-chunk Training} During inference, Simul-S2ST may face different latency requirements, and training multiple models for every latency is expensive \citep{multipath,zhang-feng-2021-universal}. To this end, we introduce multi-chunk training to improve the performance of StreamSpeech across various latency levels. In multi-chunk training, chunk size $C$ of streaming speech encoder is not fixed, but randomly sampled from a uniform distribution for $1$ to $\left|X \right|$, expressed as $C\sim \mathcal{U}\left (1,  \left|X \right|\right)$, where $c=\left|X \right|$ refers to offline S2ST. With multi-chunk training, a single StreamSpeech model can cater to different latency requirements.

\subsection{Inference}
Algorithm \ref{algorithm} illustrates the inference policy of StreamSpeech. During inference, StreamSpeech processes streaming speech inputs based on the set chunk size $C$, where each speech feature typically corresponds to 40$ms$ duration (e.g., $C=8$ means encoding speech inputs every $C\times 40=320ms$). Then StreamSpeech decodes the source tokens $\widehat{A}$ and target tokens $\widehat{Y}$ associated with the currently received speech $\widehat{X}$ through the CTC decoders for ASR and NAR-S2TT tasks. In cases where new source token is recognized, and the count of aligned target tokens surpasses the previously generated target tokens (line 5), StreamSpeech autoregressively generates the target tokens (line 7-10) and synchronously generates the corresponding units (line 11) and synthesizes the target speech (line 12); otherwise StreamSpeech waits for the next speech chunk of size $C$. Due to the proposed multi-chunk training, StreamSpeech can control the latency by adjusting chunk size $C$ during inference, where the smaller $C$ will lead to lower latency.

\begin{algorithm}[t]
\footnotesize
\caption{Inference of StreamSpeech}\label{algorithm}
  \SetKwInOut{Input}{Input}\SetKwInOut{Output}{Output}
  \Input{streaming speech inputs $X$, chunk size $C$, current received speech $\widehat{X}$}
  \Output{target speech outputs $S$}
  \While{ $|\widehat{X}|\leq \left| X\right|$}{
    generate ASR results $\widehat{A}$, with Eq.(\ref{eq:ctc_decoder});\\
    generate NAR-S2TT results $\widehat{Y}$, with Eq.(\ref{eq:ctc_decoder});\\
    \eIf(\tcp*[f]{$\!\!$WRITE}){$|\widehat{A}|\!>\!|A|$ \textbf{\textup{and}} $|\widehat{Y}|\!>\!|Y|$}{
        $A=\widehat{A}$;\\
        \While{$|Y|<|\widehat{Y}|$ \textbf{\textup{and}} $ Y_{-1}\neq \text{<eos>}$}{
            generate target token $y$, with Eq.(\ref{eq:s2t});\\
            $Y.\textrm{append}(y)$
        }
        generate units $U$ of $Y$, with Eq.(\ref{eq:t2u});\\
        $S=\text{Vocoder}(U)$;\\
        \tcp*[f]{$\!\!\!\!$ output new generated speech$\;\;\;\;\;\;\;\;\;\;$}
    }
    (\tcp*[f]{$\!\!$READ}){
      wait for next speech chunk;\\
      $\widehat{X}.\textrm{append}(X_{|\widehat{X}|:|\widehat{X}|+C})$;
    }
    }
\end{algorithm}

\section{Experiments}
\subsection{Experimental Setup}
\textbf{Datasets}\quad We conduct experiments on CVSS-C benchmark \citep{jia-etal-2022-cvss}, which is a large-scale S2ST corpus derived from the CoVoST 2 speech-to-text translation corpus \citep{wang2020covost} by synthesizing the target speech using a state-of-the-art TTS system. We evaluate StreamSpeech on CVSS-C French$\rightarrow$English (Fr$\rightarrow$En), Spanish$\rightarrow$English (Es$\rightarrow$En) and German$\rightarrow$English (De$\rightarrow$En).

\textbf{Pre-processing}\quad Following \citet{inaguma-etal-2023-unity}, we convert the source speech to 16000Hz and generate target speech with 22050Hz.
For source speech, we compute 80-dimensional mel-filterbank features \citep{Povey:192584} and apply global-level cepstral mean-variance normalization, where each speech feature corresponds to 40$ms$ duration. For target speech, we extract the discrete units via mHuBERT\footnote{\url{https://dl.fbaipublicfiles.com/hubert/mhubert_base_vp_en_es_fr_it3.pt}} \citep{popuri22_interspeech}, and synthesize target speech through a pre-trained unit-based HiFi-GAN vocoder\footnote{\url{https://dl.fbaipublicfiles.com/fairseq/speech_to_speech/vocoder/code_hifigan/mhubert_vp_en_es_fr_it3_400k_layer11_km1000_lj}} \citep{NEURIPS2020_c5d73680}. For source and target text, we use SentencePiece \citep{kudo-richardson-2018-sentencepiece} to generate a unigram vocabulary of size $6000$, respectively.

\subsection{Systems Settings}
Since StreamSpeech can be applied to simultaneous and offline S2ST, we compare StreamSpeech with offline S2ST and Simul-S2ST models.

Offline S2ST baselines include \textbf{S2UT} \citep{lee-etal-2022-direct}, \textbf{Translatotron} \citep{jia19_interspeech}, \textbf{Translatotron 2} \citep{pmlr-v162-jia22b}, \textbf{DASpeech} \citep{fang2023daspeech} and \textbf{UnitY} \citep{inaguma-etal-2023-unity}, where UnitY is the state-of-the-art offline S2ST model and is the basic framework of seamlessM4T \citep{communication2023seamlessm4t}. Refer to Appendix \ref{app:baseline} for detailed introduction to these offline baselines.

Simul-S2ST baselines include:

\textbf{Wait-k} \citep{ma-etal-2019-stacl} Wait-k policy first waits for $k \!\times\!320ms$ of speech, and then generates a target word every 320$ms$ \citep{ma-etal-2020-simulmt}. We apply wait-k policy on UnitY, where the first pass adopts wait-k policy, and then the second pass generates units until <eos> token.

\textbf{ASR+HMT+TTS (cascaded)} \citep{zhang2023hidden} Hidden Markov Transformer\footnote{\url{https://github.com/ictnlp/HMT}} (HMT) is the state-of-the-art simultaneous text-to-text translation model. We train the streaming ASR and real-time TTS model and add them before and after HMT to form a cascaded Simul-S2ST system.

\textbf{DiSeg+TTS (cascaded)} \citep{zhang-feng-2023-end} Differentiable segmentation\footnote{\url{https://github.com/ictnlp/DiSeg}} (DiSeg) is the state-of-the-art simultaneous speech-to-text translation model. We also add real-time TTS model after DiSeg to form a cascaded Simul-S2ST system.

\textbf{StreamSpeech} Our direct Simul-S2ST model.

\begin{table*}[t]
\centering
\small
\begin{tabular}{l|c|cc|cc|cc|cc} \toprule
\multirow{2}{*}{\textbf{Models}} & \multirow{2}{*}{\textbf{\#Param.}} & \multicolumn{2}{c|}{\textbf{Fr$\rightarrow$En}} & \multicolumn{2}{c|}{\textbf{Es$\rightarrow$En}} & \multicolumn{2}{c|}{\textbf{De$\rightarrow$En}}     & \multicolumn{2}{c}{\textbf{Average}} \\
                                 &                                   & greedy           & beam10         & greedy           & beam10         & greedy         & beam10        & greedy         & beam10         \\ \midrule
\textbf{Ground Truth}            & -                                 & \multicolumn{2}{c|}{84.52}          & \multicolumn{2}{c|}{88.54}          & \multicolumn{2}{c|}{75.53}       & \multicolumn{2}{c}{82.86}         \\ \midrule
\textbf{S2UT}                    & 73M                               & 20.91                & 22.23           & 16.94                & 18.53           & 2.46              & 2.99           & 13.44               & 14.58           \\
\textbf{Translatotron}           & 79M                               & 16.96                & /           & 8.72                & /            & 1.97              & /           & 9.22               & /            \\
\textbf{Translatotron 2}         & 87M                               & 25.49            & 26.07           & 22.35            & 22.93           & 16.24          & 16.91          & 21.36           & 21.97           \\
\textbf{DASpeech}                & 93M                               & 25.03                & /           & 21.37                & /                & 16.14              & /          & 20.85               & /           \\
\textbf{UnitY}                   & 67M                               & 26.90            & 27.77           & 23.93            & 24.95           & 18.19          & 18.74          & 23.01           & 23.82           \\ \midrule
\textbf{StreamSpeech}            & 70M                               & \textbf{27.58}$^{**}\!\!\!\!\!$   & \textbf{28.45}$^{**}\!\!\!\!\!$  & \textbf{26.16}$^{**}\!\!\!\!\!$   & \textbf{27.25}$^{**}\!\!\!\!\!$  & \textbf{19.72}$^{**}\!\!\!\!\!$ & \textbf{20.93}$^{**}\!\!\!\!\!$ & \textbf{24.49}  & \textbf{25.54} \\ \bottomrule
\end{tabular}
\caption{Offline S2TT results (ASR-BLEU) on CVSS-C Fr$\rightarrow$En, Es$\rightarrow$En, De$\rightarrow$En test sets. We report the results under greedy and beam=10 decoding, where Translatotron only supports greedy decoding and DASpeech uses Viterbi decoding. $^{**}$ means the improvements over the SOTA UnitY are statistically significant ($p < 0.01$).}
\label{tab:offline}
\end{table*}

All implementations are adapted from Fairseq Library \citep{ott-etal-2019-fairseq}. StreamSpeech uses basically the same settings as UnitY \citep{inaguma-etal-2023-unity}, and the introduced CTC decoder consists of only a fully connected layer. Other model configurations and training details are reported in Appendix \ref{app:config}. The only hyperparameter that needs to be set in StreamSpeech is the upsampling rate $r$ in NAR T2U generation, where we set $r=25$ based on validation in Appendix \ref{app:r}. For cascaded systems, the streaming ASR and real-time TTS modules use the streaming encoder and non-autoregressive text-to-unit module, identical to those used in StreamSpeech, for a fair comparison.

\subsection{Evaluation}
We apply SimulEval\footnote{\url{https://github.com/facebookresearch/SimulEval}} \citep{ma-etal-2020-simuleval} to evaluate the Simul-S2ST from both quality and latency.

\textbf{Quality}\quad We evaluate S2ST quality using ASR-BLEU toolkit\footnote{\url{https://github.com/facebookresearch/fairseq/tree/ust/examples/speech_to_speech/asr_bleu}}, which first transcribes the translated speech into text using a pre-trained ASR model and then calculates the SacreBLEU \citep{post-2018-call} score with reference. We also use BLASER 2.0 to assess the generated speech's quality and the results are reported in Appendix \ref{sec:blaser} and \ref{sec:numerical}.

\begin{table}[t]
\centering\small
\begin{tabular}{L{1.6cm}C{0.5cm}C{0.55cm}C{0.5cm}C{0.55cm}C{0.5cm}C{0.55cm}}\toprule
\multirow{2}{*}{\textbf{Models}} & \multicolumn{2}{c}{\textbf{Fr$\rightarrow$En}}                                  & \multicolumn{2}{c}{\textbf{Es$\rightarrow$En}}                                  & \multicolumn{2}{c}{\textbf{De$\rightarrow$En}}                                  \\ \cmidrule(lr){2-3}\cmidrule(lr){4-5}\cmidrule(lr){6-7}
                                 &\scriptsize \begin{tabular}[c]{@{}c@{}}$\!$ASR-\\ $\!$BLEU\end{tabular} &\scriptsize $\!\!\!$Speedup       & \scriptsize \begin{tabular}[c]{@{}c@{}}$\!$ASR-\\ $\!$BLEU\end{tabular} & \scriptsize $\!\!\!$Speedup       & \scriptsize \begin{tabular}[c]{@{}c@{}}$\!$ASR-\\ $\!$BLEU\end{tabular} &\scriptsize $\!\!\!$Speedup       \\\midrule
\textbf{UnitY}                   & 27.77                                               & 1.0×          & 24.95                                               & 1.0×          & 18.74                                               & 1.0×          \\
\textbf{StreamSpeech}            & \textbf{28.45}                                      & \textbf{3.6×} & \textbf{27.25}                                      & \textbf{4.5×} & \textbf{20.93}                                      & \textbf{4.5×} \\\bottomrule
\end{tabular}
\caption{Speedup of StreamSpeech.}
\label{tab:speed}
\end{table}

\textbf{Latency}\quad We use Average Lagging (AL) \citep{ma-etal-2020-simulmt} to evaluate the latency, where AL measures the average duration ($ms$) that outputs lag behind inputs. We also measure the computation-aware latency, which includes the computational time of the model. The computation-aware latency is evaluated on 1 NVIDIA RTX 3090 GPU with batch-size$=$1. More latency metrics are reported in Appendix \ref{app:metric} to show latency performance.

\subsection{Main Results}
We conduct experiments in both offline S2ST and Simul-S2ST tasks.

\textbf{Offline S2ST}\quad Table \ref{tab:offline} reports the performance of StreamSpeech in offline S2ST, where StreamSpeech outperforms the state-of-the-art UnitY with an average improvement of 1.5 BLEU. StreamSpeech uses two-pass architecture and achieves significant improvements over S2UT and Translatotron, which use one-pass architecture. For two-pass architecture, DASpeech employs NAR architecture in both first and second passes \citep{fang2023daspeech}, while UnitY uses AR architecture in two passes \citep{inaguma-etal-2023-unity}. StreamSpeech uses AR architecture in S2TT task (first pass) that involves more reordering and context dependence, and NAR architecture in T2U task (second pass) that is basically monotonically aligned. This effectively mitigates the impact of the NAR architecture on modeling capabilities and meanwhile captures the alignment between text and unit. Overall, multi-task learning not only guides the policy, but also provides intermediate supervision for translation, yielding superior offline S2ST performance.  

\begin{figure*}[t]
\centering
\subfigure[Fr$\rightarrow$En]{
\includegraphics[width=0.315\textwidth]{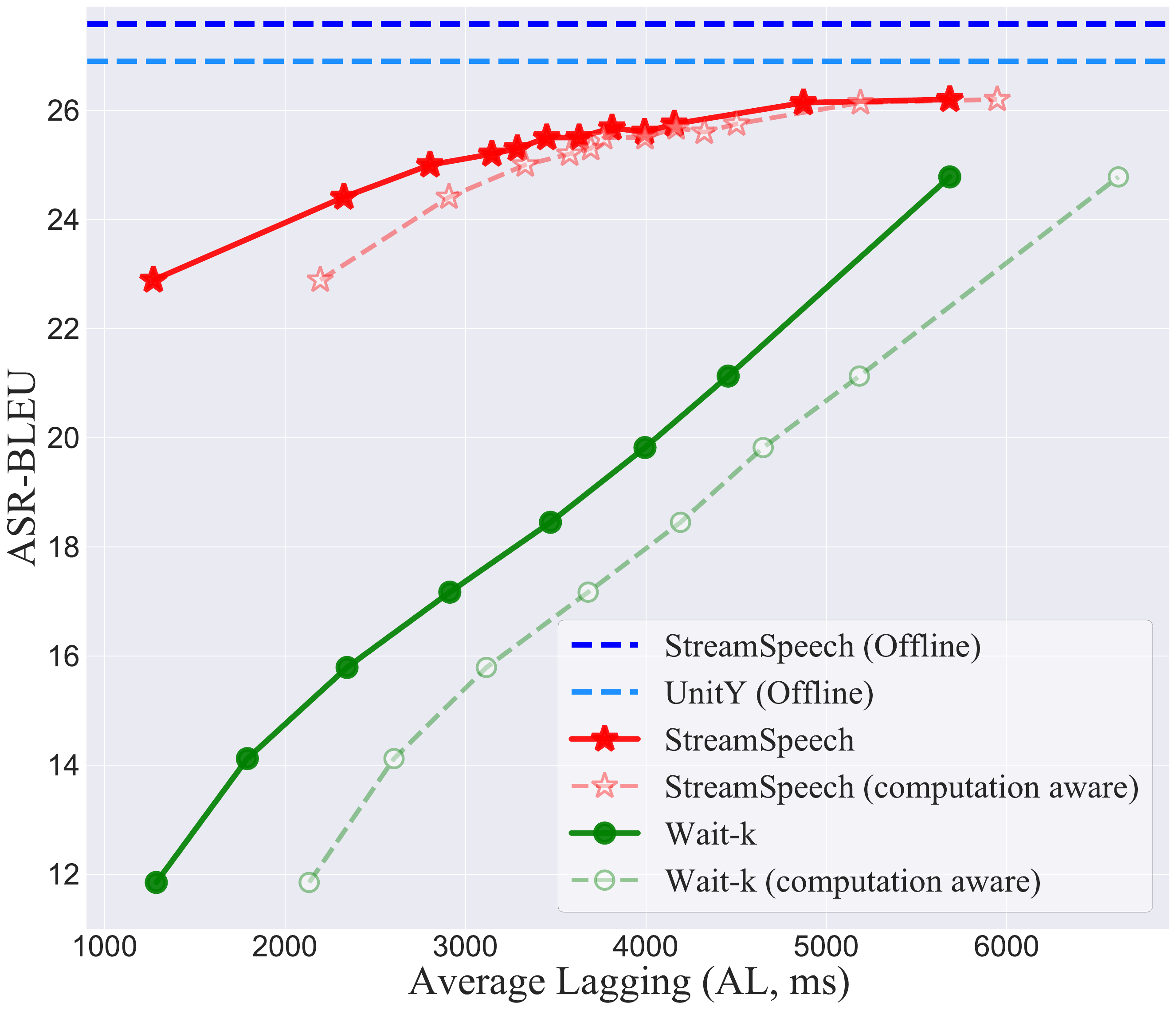}\label{fig:main1}
}
\subfigure[Es$\rightarrow$En]{
\includegraphics[width=0.315\textwidth]{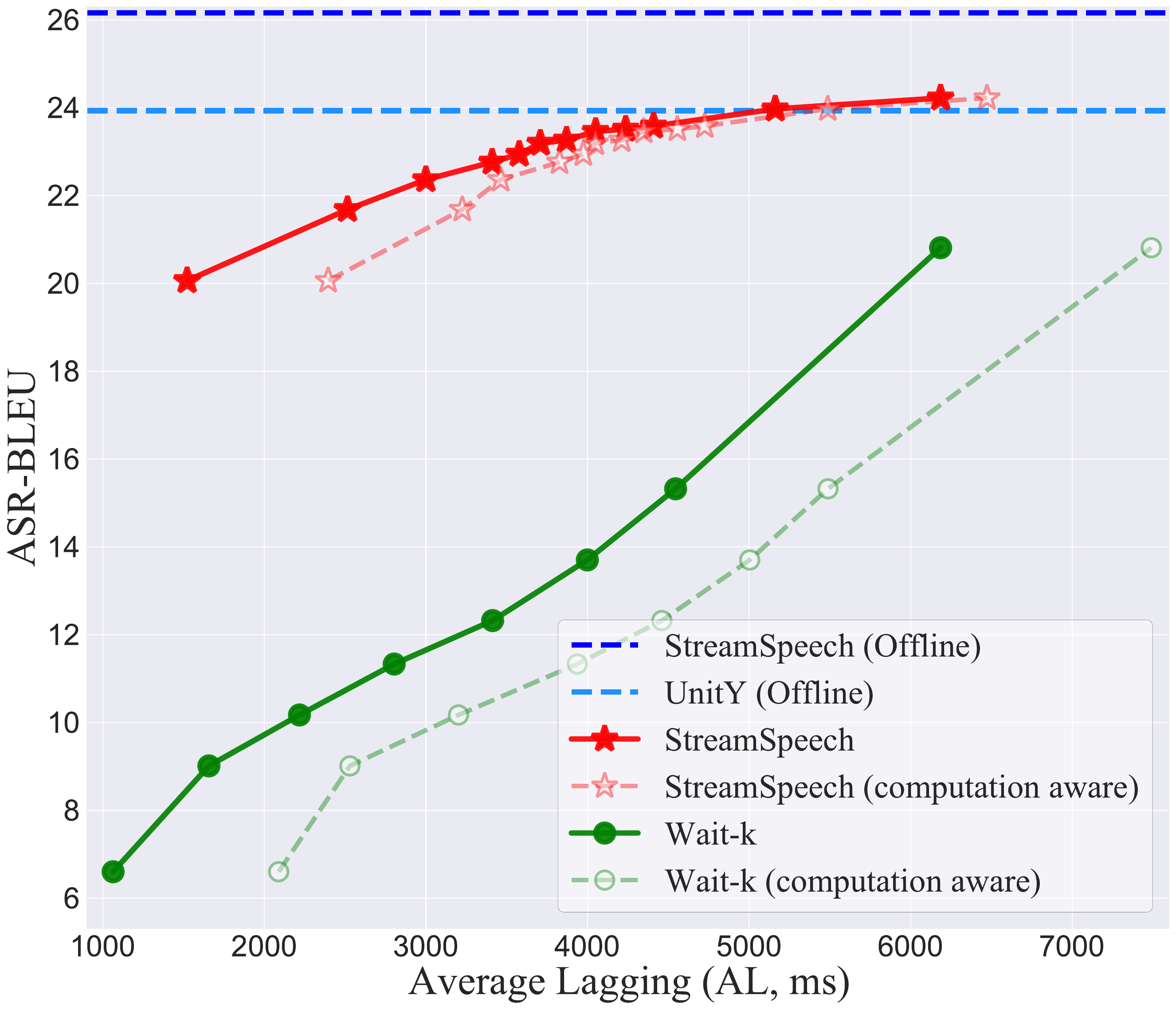}\label{fig:main2}
}
\subfigure[De$\rightarrow$En]{
\includegraphics[width=0.315\textwidth]{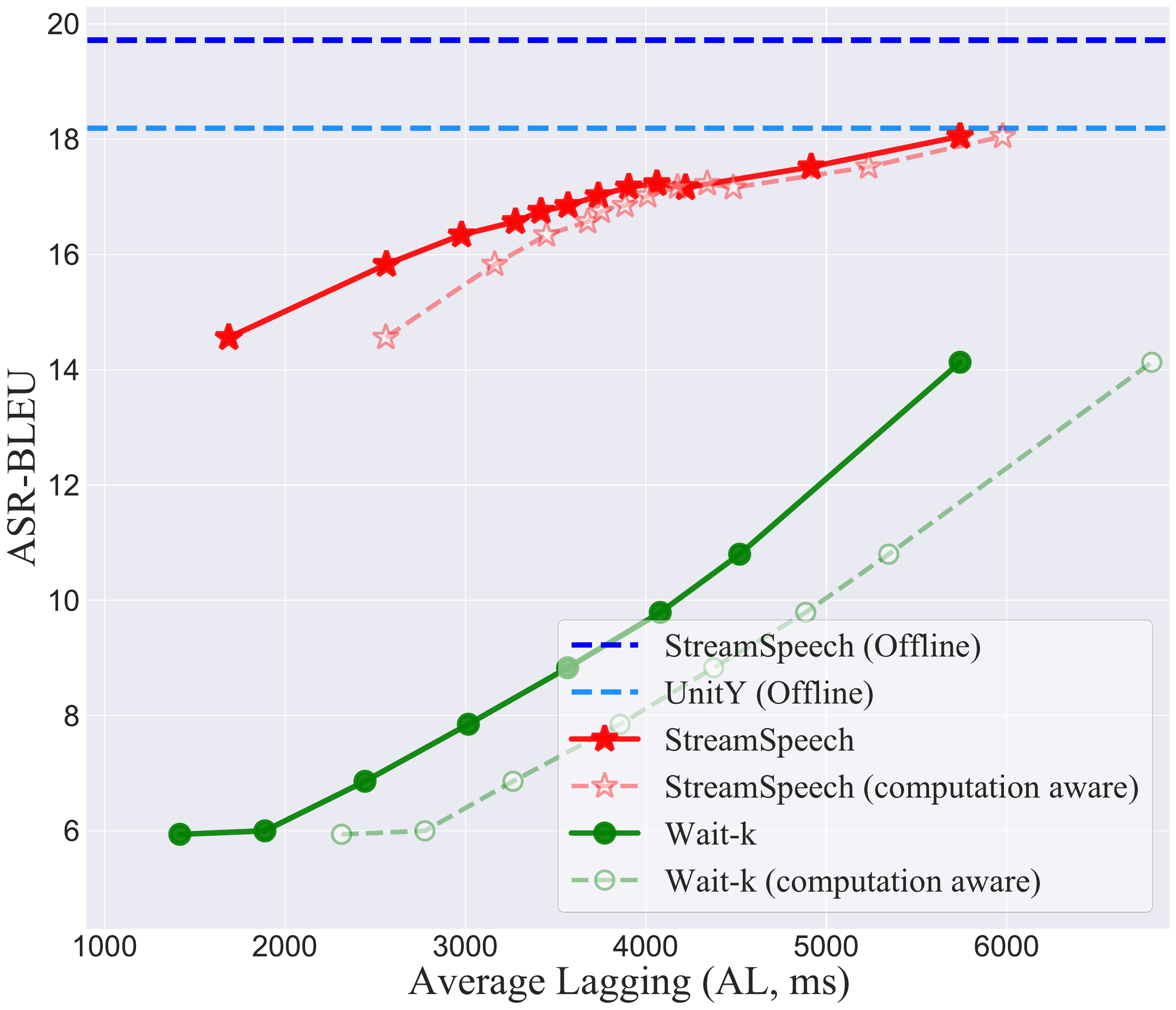}\label{fig:main3}
}

\caption{Simul-S2ST results (quality against latency) on CVSS-C Fr$\rightarrow$En, Es$\rightarrow$En, De$\rightarrow$En test sets. The hollow points represent computation-aware latency, which includes the inference time consumed by the model. Some simultaneous outputs of StreamSpeech can be heard at \href{https://ictnlp.github.io/StreamSpeech-site/}{https://ictnlp.github.io/StreamSpeech-site/}.}
\label{fig:main}
\end{figure*}

\begin{table*}[t]
\centering
\small
\begin{tabular}{l|C{0.5cm}C{1.2cm}C{1.0cm}C{0.6cm}|ccccc} \toprule
\multirow{2}{*}{\textbf{Models}}       & \multicolumn{4}{c|}{\textbf{Tasks}}    & \textbf{ASR}   & \textbf{NAR-S2TT} & \textbf{AR-S2TT} & \textbf{S2UT}   & \textbf{S2ST}     \\
                              & \scriptsize ASR & \scriptsize NAR-S2TT &\scriptsize AR-S2TT & \scriptsize S2UT & \scriptsize WER$\downarrow$   & \scriptsize BLEU$\uparrow$$\;\;$ ACC$\uparrow$     & \scriptsize BLEU$\uparrow$$\;\;$ ACC$\uparrow$   & \scriptsize  BLEU$\uparrow$  & \scriptsize ASR-BLEU$\uparrow$ \\ \midrule
\textbf{UnitY}                         & \xmark   & \xmark       & \cmark      & \cmark   & /     & /        & 31.31$\;\;\;$61.0   & \textbf{33.47}  & 27.77    \\ \midrule
\multirow{4}{*}{\textbf{StreamSpeech}} & \xmark   & \xmark       & \cmark      & \cmark   & /     & /        & 31.20$\;\;\;$61.5   & 31.37  & 27.47    \\
                              & \xmark   & \cmark       & \cmark      & \cmark   & /     & 22.95$\;\;\;$59.9    & 31.56$\;\;\;$61.1   & 31.15 & 27.73    \\
                              & \cmark   & \xmark       & \cmark      & \cmark   & 20.70 & /        & 32.28$\;\;\;$62.3   & 31.42  & 28.18    \\
                              & \cmark   & \cmark       & \cmark      & \cmark   & \textbf{20.55} & \textbf{23.82}$\;\;\;$\textbf{60.9}    & \textbf{32.60}$\;\;\;$\textbf{62.4}   & 31.72  & \textbf{28.45}   \\\bottomrule
\end{tabular}
\caption{Ablation study of multi-task learning on offline S2ST, evaluated on CVSS-C Fr$\rightarrow$En test set. We report word error rate (WER) for ASR task, BLEU score and 1-gram accuracy (ACC) for NAR-S2TT and AR-S2TT tasks, BLEU score (computes on unit sequences) for S2UT task, and ASR-BLEU score for S2ST task.}
\label{tab:multitask}
\end{table*}

\textbf{Speedup of StreamSpeech}\quad To explore the inference efficiency of StreamSpeech, we report the speedup of StreamSpeech compared to UnitY in Table \ref{tab:speed}. In the two-pass architecture, StreamSpeech employs an autoregressive structure in the first pass for translation and a non-autoregressive structure in the second pass for speech synthesis (where the sequences are longer but monotonically aligned). This AR+NAR two-pass architecture brings significant speedup while maintaining translation quality.

\textbf{Simul-S2ST}\quad Figure \ref{fig:main} shows the Simul-S2ST performance of StreamSpeech, where StreamSpeech outperforms Wait-k under all latency, particularly exhibiting a roughly 10 BLEU improvement under low latency. Wait-k stands as the most widely used policy and achieves good performance on simultaneous T2TT and S2TT \citep{ma-etal-2019-stacl,ma-etal-2020-simulmt}. For the Simul-S2ST task where the source and target sequences are both continuous speech, StreamSpeech's policy derived from alignments enables the model to translate at more appropriate moments and generate coherent target speech, resulting in significant advantages. Moreover, concerning computation-aware latency, StreamSpeech introduces only a marginal increase in parameters, thus avoiding notable inference overhead.

\textbf{Direct Simul-S2ST v.s. Cascaded Simul-S2ST}\quad Figure \ref{fig:cascade} presents a comparison between the direct and cascaded Simul-S2ST models, evaluated on the CVSS-C Fr$\rightarrow$En test set. The results suggest a general superiority of the direct model over the cascaded systems. Specifically, when we decompose the direct StreamSpeech into two modules ``S2TT+TTS'', error accumulation leads to a 1 BLEU decrease under low latency, even with the same policy. Furthermore, compared to the cascaded system comprising state-of-the-art HMT and DiSeg, StreamSpeech demonstrates a significant advantage, underscoring the superiority of direct StreamSpeech in Simul-S2ST task.

\section{Analyses}

\subsection{Effect of Multi-task Learning}

StreamSpeech jointly optimizes S2UT, AR-S2TT, ASR, and NAR-S2TT tasks through multi-task learning. To verify the effect of multi-task learning, we conduct an ablation study of auxiliary tasks on CVSS-C Fr$\rightarrow$En offline S2ST. 

As reported in Table \ref{tab:multitask}, the introduction of auxiliary tasks effectively improves the performance of S2ST. Multi-task learning offers staged intermediate supervision for each module within StreamSpeech \citep{tang-etal-2021-improving,9415058}, thereby enhancing overall performance. 
Furthermore, it is noteworthy that NAR-S2TT exhibits a notable gap in BLEU scores compared to AR-S2TT, while the 1-gram accuracy shows minimal differences. This highlights the rationale behind utilizing NAR-S2TT for aligning source speech and target text and employing AR-S2TT for translation.

\begin{figure*}[t]
\begin{minipage}[b]{0.32\textwidth}
    \centering
    \includegraphics[width=\textwidth]{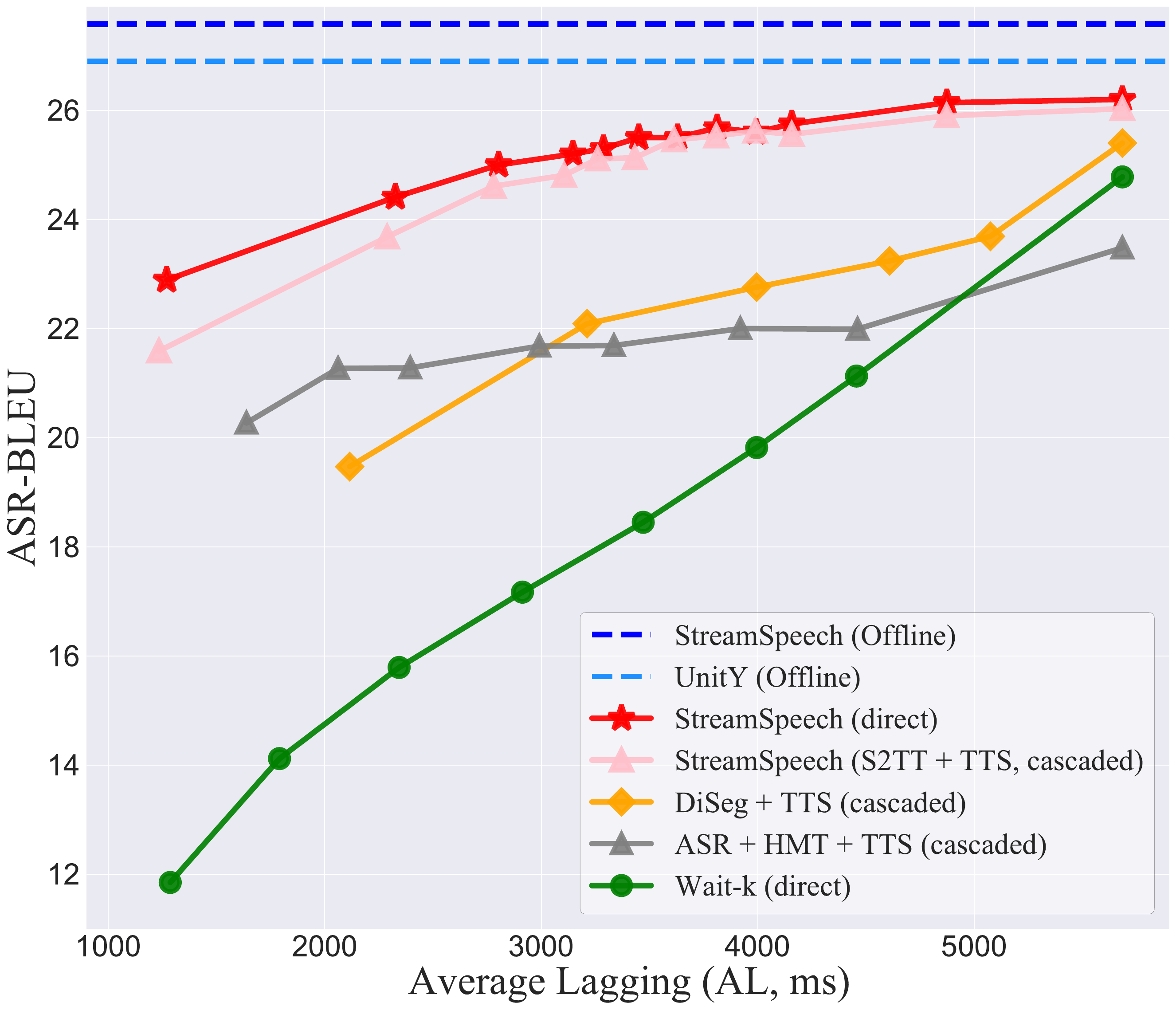}
    \caption{Comparison of direct and cascaded Simul-S2ST systems.}
    \label{fig:cascade}
  \end{minipage}
  \quad
  \begin{minipage}[b]{0.32\textwidth}
    \centering
    \includegraphics[width=\textwidth]{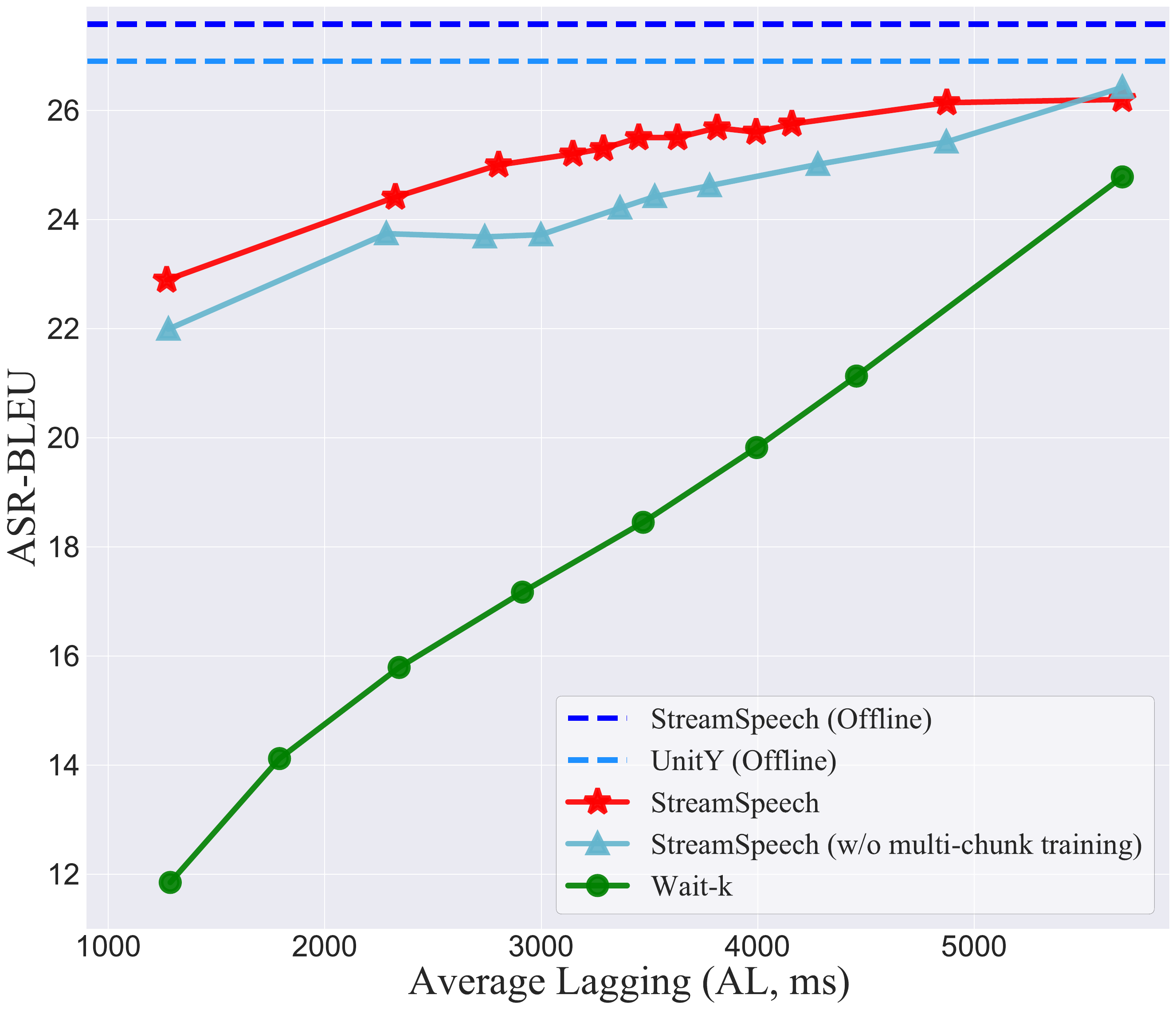}
    \caption{Effect of multi-chunk training in StreamSpeech.}
    \label{fig:ab_multichunk}
  \end{minipage}
  \quad
  \begin{minipage}[b]{0.32\textwidth}
    \centering
    \includegraphics[width=\textwidth]{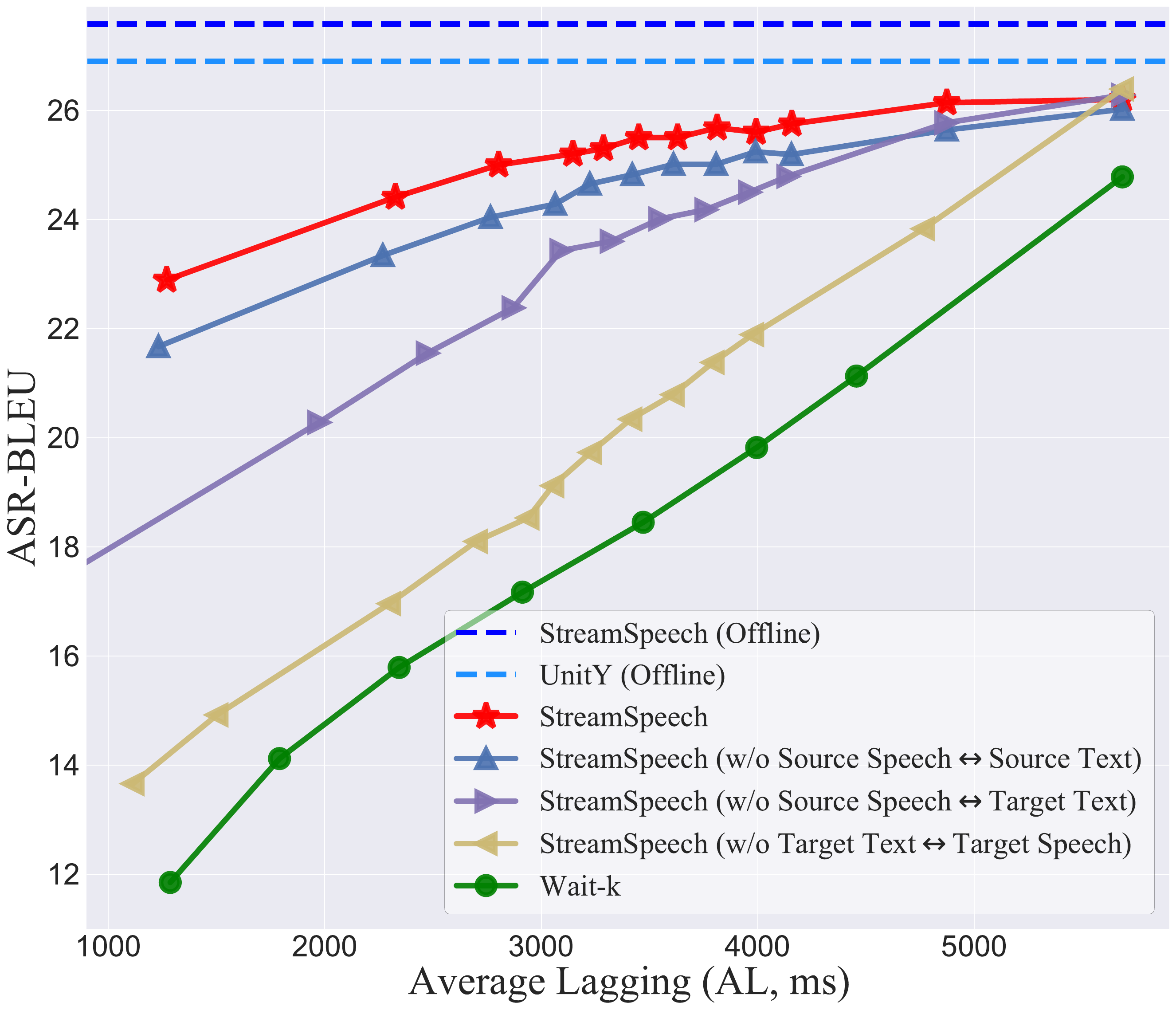}
    \caption{Ablation study on alignments in policy.}
    \label{fig:ab_alignment}
  \end{minipage}
\end{figure*}

\subsection{Superiority of Multi-chunk Training}

\begin{table}[t]
\small
\begin{tabular}{c|ccccc} \toprule
{Train} \textbackslash{} {Test} & $\!\!\!C\!\!=\!\!8\!\!\!$              & $\!\!\!C\!\!=\!\!16\!\!\!$             & $\!\!\!C\!\!=\!\!32\!\!\!$             & $\!\!\!C\!\!=\!\!64\!\!\!$             & $\!\!\!C\!\!=\!\!\infty\!\!\!$                  \\ \midrule
$C=8\;\:$                         & {24.91}    & 24.72          & 25.03          & 24.82          & 23.37                \\
$C=16$                        & 24.18          & {25.64}    & 25.75          & 25.62          & 24.76                \\
$C=32$                        & 23.06          & 24.69          & {25.82}    & 25.85          & 25.75                \\
$C=64$                        & 19.55          & 22.77          & 24.63          & {25.94}    & 26.41                \\ 
$C=\infty$                       & 1.42           & 7.12           & 14.58          & 21.76          & {\textbf{26.90}} \\ \midrule
Multi-Chunk               & \textbf{25.34} & \textbf{25.97} & \textbf{26.31} & \textbf{26.61} & 26.47             \\ \bottomrule   
\end{tabular}
\caption{Offline S2ST results on various chunk size $C$ of streaming encoder during training and testing.}
\label{tab:multichunk}
\end{table}

To enhance the performance of StreamSpeech under various latency, we propose multi-chunk training. Figure \ref{fig:ab_multichunk} illustrates the Simul-S2ST performance on Fr$\rightarrow$En test set when employing multi-chunk training and training multiple separate models for different latency. The results indicate that multi-chunk training performs better under all latency. More importantly, multi-chunk training enables StreamSpeech to handle Simul-S2ST under various latency conditions using just one model.

\textbf{Chunk-based Conformer}\quad To further evaluate the impact of multi-chunk training on the modeling capabilities of chunk-based Conformer, we conduct experiments by training StreamSpeech with various chunk sizes $C$ and testing them with different test chunk sizes in offline S2ST. The results are reported in Table \ref{tab:multichunk}, evaluated on CVSS-C Fr$\rightarrow$En test set. The results indicate that models trained with a single chunk size often excel only at a particular test chunk size and struggle to adapt to others. Multi-chunk training equips StreamSpeech with the capability to adapt to different chunk sizes, thereby enabling it to handle S2ST under various latency conditions using a single model. Notably, multi-chunk training also demonstrates superior performance at smaller chunk sizes, which in line with previous findings suggesting that incorporating future information during training has a positive improvement \citep{ma-etal-2019-stacl,future-guided,laf,guo2023glancing}.

\subsection{Analysis on StreamSpeech Policy}
\label{sec:policy_ab}

StreamSpeech models alignments between source speech and source/target text, target text and target speech, thereby allowing for the adaptive decision of READ/WRITE actions. To assess the significance of these three alignments, we present the CVSS-C Fr$\rightarrow$En performance when removing one of them (refer to Appendix \ref{app:ab} for detailed introduction of ablation settings) in Figure \ref{fig:ab_alignment}. 

The results underscore the pivotal role of modeling the alignment between target text and target speech through NAR text-to-unit module, as the number of units corresponding to text is often diverse, and the proposed unit CTC decoder effectively addresses this issue. Besides, the alignment between source speech and source/target text ensures StreamSpeech starts translating at appropriate moments and generates a reasonable number of target tokens, where removing any of these alignment components results in performance degradation. Totally, the alignments involved in StreamSpeech are reasonable and can be jointly trained with translation through multi-task learning.

\subsection{Performance on Auxiliary Tasks: Streaming ASR and Simultaneous S2TT}

\begin{table}[t]
\centering\small
\begin{tabular}{lccc} \toprule
\textbf{Models}                        & \textbf{\#Parm.}                                                            & \textbf{AL ($ms$)$\downarrow$} & \textbf{WER$\downarrow$} \\ \midrule
\textbf{Wav2Vec2-large}                & 315M                                                                        & 5684.38     & 26.17        \\
\textbf{Whisper-base}                  & 74M                                                                        & 5684.38     & 38.04        \\\midrule
\multirow{4}{*}{\textbf{StreamSpeech}} & \multirow{4}{*}{\begin{tabular}[c]{@{}c@{}}70M\\ (33M used)\end{tabular}} & 109.127     & 25.46        \\
                                       &                                                                             & 267.891     & 25.54        \\
                                       &                                                                             & 431.652     & 25.20        \\
                                       &                                                                             & 757.989     & 24.67  \\   \bottomrule
\end{tabular}
\caption{Streaming ASR results on Fr$\rightarrow$En test set.}
\label{fig:streaming_asr}  
\end{table}

\begin{figure}[t]
    \centering
    \includegraphics[width=0.35\textwidth]{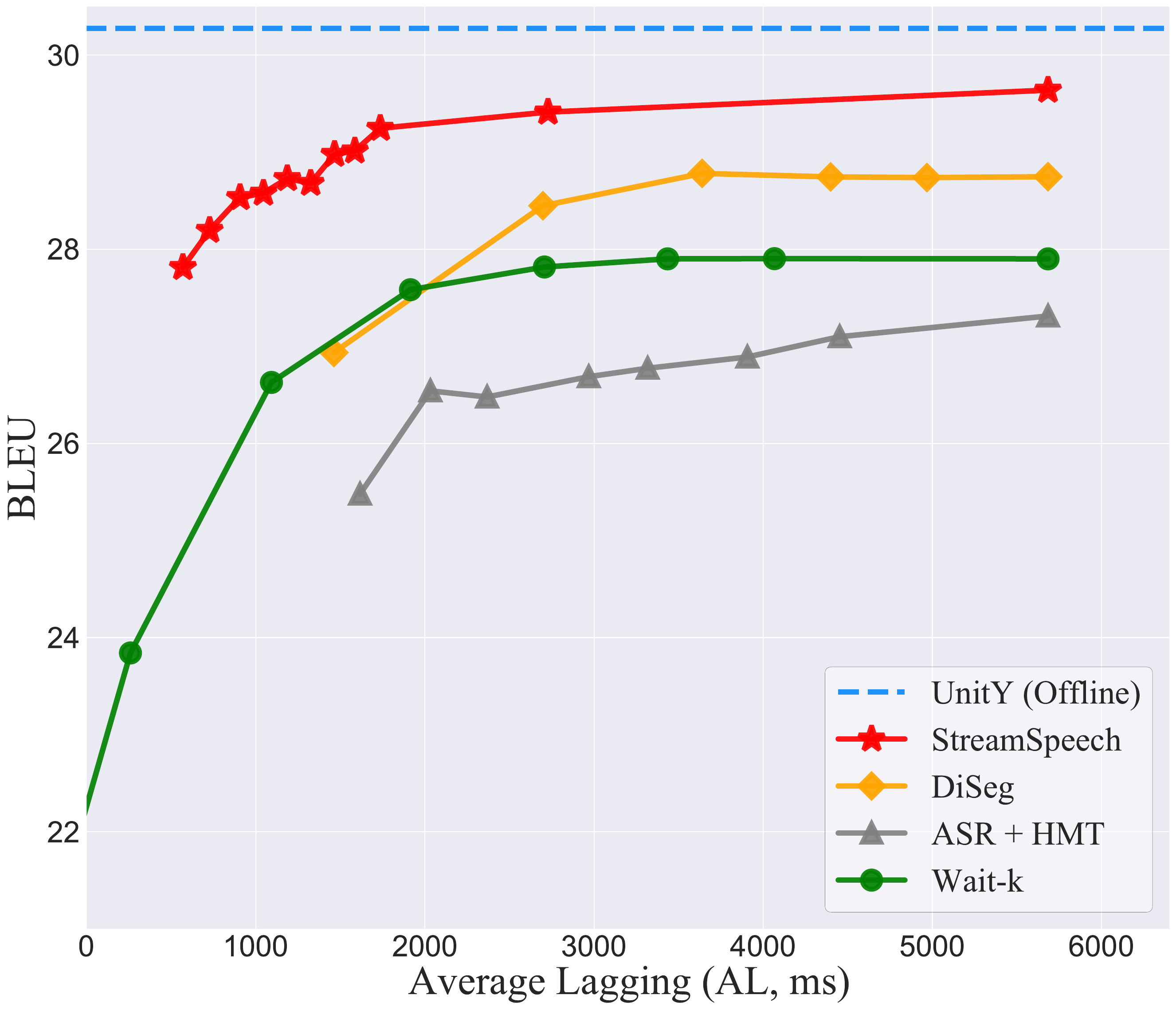}
    \caption{Simultaneous speech-to-text translation results on Fr$\rightarrow$En test set.}
    \label{fig:simul-s2tt}
\end{figure}

StreamSpeech jointly learns translation and policy through multi-task learning. As an additional product, StreamSpeech can output intermediate results of ASR and S2TT, offering users a more comprehensive experience. To evaluate StreamSpeech's performance on these auxiliary tasks, we present the results on the streaming ASR and Simul-S2TT tasks in Table \ref{fig:streaming_asr} and Figure \ref{fig:simul-s2tt}, respectively, assessed on the CVSS-C Fr$\rightarrow$En test set.

For streaming ASR, StreamSpeech achieves performance comparable to Wav2Vec2-large \citep{NEURIPS2020_92d1e1eb} and Whisper-base \citep{radford2022robust} with an average lagging of 100$ms$. For Simul-S2TT, StreamSpeech can generate high-quality translation with an average lagging of 2000$ms$. Overall, StreamSpeech excels in delivering high-quality Simul-S2ST while also providing intermediate results to users as additional references. It's important to note that although intermediate results can be presented, StreamSpeech does not utilize them during inference, but uses hidden states to connect each module, making StreamSpeech a direct Simul-S2ST model.

\section{Related Work}

Recent research often focuses on simultaneous text-to-text (Simul-T2TT) and speech-to-text (Simul-S2TT) translation.

\textbf{Simul-T2TT}\quad Simul-T2TT methods fall into fixed and adaptive. For fixed methods, \citet {ma-etal-2019-stacl} proposed wait-k policy, which waits for $k$ tokens before alternately READ/WRITE one token \citep{multipath,zheng-etal-2020-simultaneous,zhang-feng-2021-icts,zhang-feng-2021-universal}. For adaptive methods, monotonic attention \citep{Arivazhagan2019,Ma2019a,dualpath,gma}, alignments \citep{zhang2023hidden,guo-etal-2023-learning}, non-autoregressive architecture \citep{ma-etal-2023-non,nast} or language models \citep{dst,guo2024sillm,zhang2023bayling} are employed to dynamically perform Simul-T2TT. On top of these policies, some training methods are proposed to enhance the performance of policy \citep{laf,wait-info,post-eval,guo-etal-2023-simultaneous}.

\textbf{Simul-S2TT}\quad Simul-S2TT methods focus on the segmentation of speech. \citet{ma-etal-2020-simulmt} proposed fixed pre-decision to divide speech into equal-length segments. Some adaptive methods split the speech inputs into words or segments \citep{ren-etal-2020-simulspeech,zeng-etal-2021-realtrans,chen-etal-2021-direct,dong-etal-2022-learning,ITST,zhang-etal-2022-learning,zhang-feng-2023-end,zhang2023unified}, and then apply Simul-T2TT policy. Other methods apply offline model to Simul-S2TT \citep{papi-etal-2023-attention,fu-etal-2023-adapting,dugan2023learning}.

We introduce StreamSpeech, an ``All in One'' seamless model for offline and simultaneous ASR, translation and synthesis. Compared with SOTA SeamlessStreaming (based on UnitY architecture) \citep{communication2023seamless}, StreamSpeech does not design any additional simultaneous policy such as EMMA, but directly jointly learns translation and policy via multi-task learning.

\section{Conclusion}

In this paper, we propose StreamSpeech, an ``All in One'' seamless model that handles streaming ASR, simultaneous translation and real-time speech synthesis via a unified model. Experiments show the superiority of StreamSpeech on offline S2ST, streaming ASR, simultaneous S2TT and simultaneous S2ST. Moreover, intermediate products such as ASR or S2TT results can also be presented to user during translation process as a reference, offering a better low-latency communication experience.

\section*{Acknowledgements}
We thank all the anonymous reviewers for their insightful and valuable comments. This work was supported by a grant from the National Natural Science Foundation of China (No. 62376260).

\section*{Limitations}

In this paper, we propose StreamSpeech, a direct Simul-S2ST model that jointly learns translation and policy in a unified framework of multi-task learning. StreamSpeech can achieve high-quality speech-to-speech translation with low latency. However, StreamSpeech currently focuses on synthesizing target speech with a unified voice, which limits its ability to clone the source speech's voice characteristics. Given that voice cloning can enhance the authenticity of low-latency communication, we will explore integrating voice cloning capabilities into StreamSpeech as part of future work.

\bibliography{custom}
\clearpage

\appendix

\section{Calculation of Expected Token Number in CTC sequence}
\label{app:ctc}

StreamSpeech leverages ASR and NAR-S2TT tasks to learn the alignment between the source speech and the source/target text, and then makes READ/WRITE decisions based on the token number corresponding to the current received speech. Since the alignments are obtained through the CTC decoder, which involves blank and repeated tokens, we need to count the number of tokens that can be decoded by the CTC sequence. During inference, it is straightforward to remove duplicates and blanks from the CTC sequence to count the tokens corresponding to the speech. However, during training, we calculate the expected number of tokens corresponding to the CTC sequence.

For the excepted number of source tokens aligned to the speech inputs $X$, $\mathcal{N}^{asr}_{j}$ is calculated as: 
\begin{gather}
\begin{aligned}
    \mathcal{N}^{asr}_{j}=&\sum_{m=1}^{j} \Bigg( 1-p\left (\phi  \mid D^{asr}_{m}\right ) - \\
    \sum_{v\in\mathcal{V}}&\Big( p\left (v  \mid D^{asr}_{m}\right )\times p\left (v  \mid D^{asr}_{m-1}\right ) \Big)  \Bigg).
\end{aligned}
\end{gather}
where $\mathcal{N}^{asr}_{j}$ is the number of source tokens that align to $X_{\leq j}$. $\mathcal{N}^{asr}_{j}$ are calculated in an expectation manner, where $p\left (\phi  |D^{asr}_{m}\right )$ refers to generating blank token and $\sum_{v\in\mathcal{V}}( p\left (v  |D^{asr}_{m}\right )\times p\left (v  |D^{asr}_{m-1}\right ))$ refers to generating repetitive tokens over the vocabulary $\mathcal{V}$. Similarly, the number of target tokens that align to $X_{\leq j}$ is calculated in the same way, denoted as $\mathcal{N}^{nar\text{-}s2tt}_{j}$. In particular, the probabilities within the CTC sequence often tend to be discretized, resulting in minimal differences between the token counts in training and inference.

\section{Details of Ablation Study on Policy }
\label{app:ab}

In Sec.\ref{sec:policy_ab}, we conduct an ablation study on the three alignments involved in StreamSpeech, including source speech and source text, source speech and target text, target text and target speech. Here, we provide detailed explanations of the ablation study settings. 

When removing the alignment between the source speech and source text, StreamSpeech is no longer wait for recognition of new tokens corresponding to the source speech but directly controls READ/WRITE based on the number of target tokens corresponding to the source speech. When removing the alignment between the source speech and target text, StreamSpeech generates one target token once recognizing one source token. When removing the alignment between target text and target speech captured by the NAR T2U module, the second pass of StreamSpeech adopts the same autoregressive architecture as UnitY and generates units autoregressively until <eos>.

\begin{table}[t]
\centering
\small
\begin{tabular}{l|cccc} \toprule
\textbf{StreamSpeech} & $r=15$  & $r=20$  & $r=25$           & $r=30$   \\ \midrule
\textbf{BLEU of units}    & 30.48 & 31.08 & \textbf{32.00} & 31.55  \\
\textbf{ASR-BLEU}     & 26.72 & 26.75 & \textbf{28.48} & 27.91  \\\bottomrule
\end{tabular}
\caption{Performance of various upsampling rate $r$ on CVSS-C Fr$\rightarrow$En validation set. ``BLEU of units'': BLEU score computed on generated units sequence.}
\label{tab:r}
\end{table}

\section{Detailed Introduction of Baselines}
\label{app:baseline}
Here, we give a detailed introduction to offline S2ST baselines.

\textbf{S2UT} \citep{lee-etal-2022-direct} Speech-to-unit translation (S2UT) directly translates the source speech to the target unit via one-pass architecture.

\textbf{Translatotron} \citep{jia19_interspeech} Translatotron translates the source speech to the target mel-spectrogram via one-pass architecture.

\textbf{Translatotron 2} \citep{pmlr-v162-jia22b} Translatotron 2 employs a two-pass architecture that generates phonemes and mel-spectrogram successively.

\textbf{DASpeech} \citep{fang2023daspeech} DASpeech employs a two-pass architecture, where first performs non-autoregressive translation and then generates mel-spectrogram via fastspeech 2.

\textbf{UnitY} \citep{inaguma-etal-2023-unity} UnitY is the state-of-the-art S2ST model, where both first and second passes apply autoregressive encoder-decoder to generate target text and units successively.

For the cascaded Simul-S2ST system, we employ state-of-the-art methods, HMT from Simul-T2TT and DiSeg from Simul-S2TT, in conjunction with streaming ASR and real-time TTS module to accomplish Simul-S2ST.

\textbf{ASR+HMT+TTS} \citep{zhang2023hidden} Hidden Markov Transformer (HMT), which uses a hidden Markov model to correspond source tokens with the target tokens, thereby learning the optimal translating moments for generating each target token. 

\textbf{DiSeg+TTS} \citep{zhang-feng-2023-end} DiSeg learns the speech segmentation from the underlying translation model via the differentiable segmentation, and then apply wait-k policy based on the number of speech segments.

\section{Upsampling Rate in NAR T2U Generation}
\label{app:r}

The only hyperparameter that needs to be set in StreamSpeech is the upsampling rate $r$ in NAR T2U generation. Table \ref{tab:r} reports the offline S2ST performance with different $r$ on the CVSS-C Fr$\rightarrow$En validation set, with $r=25$ achieving the best performance. This finding is consistent with previous conclusions in non-autoregressive translation (NAT), where an upsampling rate of 2-3 times yielded the best performance \citep{saharia-etal-2020-non}. Therefore, we set $r=25$ in our experiments accordingly. The unit sequence length is approximately 10 times that of the subword sequence length, and with a 2-3 times upsampling rate, an overall upsampling rate of around 25 times from text sequence to unit sequence is optimal.

When training StreamSpeech for a new language, it is recommended to first estimate the length ratio between the unit sequence and the subword sequence, and then multiply this ratio by 2-3 times to determine the appropriate upsampling rate.

\section{Evaluation with BLASER 2.0}
\label{sec:blaser}
\begin{table*}[t]
\centering\small
\begin{tabular}{lcccccccc} \toprule
\multirow{2}{*}{\textbf{Models}} & \multicolumn{2}{c}{\textbf{Fr$\rightarrow$En}} & \multicolumn{2}{c}{\textbf{Es$\rightarrow$En}} & \multicolumn{2}{c}{\textbf{De$\rightarrow$En}} & \multicolumn{2}{c}{\textbf{Average}} \\
                                 & greedy           & beam10          & greedy           & beam10          & greedy           & beam10          & greedy            & beam10           \\ \midrule
\multicolumn{9}{c}{\textit{ASR-BLEU}}                                                                                                                                                  \\ \midrule
\textbf{UnitY}                   & 26.90            & 27.77           & 23.93            & 24.95           & 18.19            & 18.74           & 23.01             & 23.82            \\
\textbf{StreamSpeech}            & \textbf{27.58}            & \textbf{28.45}           & \textbf{26.16}            & \textbf{27.25}           & \textbf{19.72}            & \textbf{20.93}           & \textbf{24.49}             & \textbf{25.54}            \\\midrule
\multicolumn{9}{c}{\textit{BLASER 2.0 (Unsupervised)}}                                                                                                                                 \\\midrule
\textbf{UnitY}                   & 0.4467           & 0.4473          & 0.5090           & 0.5116          & 0.4431           & 0.4435          & 0.4663            & 0.4674           \\
\textbf{StreamSpeech}            & \textbf{0.4486}           & \textbf{0.4491}          & \textbf{0.5155}           & \textbf{0.5178}          &\textbf{ 0.4514}           & \textbf{0.4544}          & \textbf{0.4719}            & \textbf{0.4738}           \\\midrule
\multicolumn{9}{c}{\textit{BLASER 2.0 (QE)}}                                                                                                                                           \\\midrule
\textbf{UnitY}                   & 3.1674           & 3.1772          & 3.3020           & 3.3278          & 3.1322           & 3.1537          & 3.2006            & 3.2195           \\
\textbf{StreamSpeech}            & \textbf{3.1779}           & \textbf{3.1872}          & \textbf{3.3442}           & \textbf{3.3669}          &\textbf{ 3.1698}           & \textbf{3.2033}          & \textbf{3.2307 }           & \textbf{3.2525 }          \\\midrule
\multicolumn{9}{c}{\textit{BLASER 2.0 (Ref)}}                                                                                                                                          \\\midrule
\textbf{UnitY}                   & 3.1744           & 3.1965          & 3.2213           & 3.2638          & 2.9125           & 2.9372          & 3.1028            & 3.1325           \\
\textbf{StreamSpeech}            & \textbf{3.1989}           & \textbf{3.2200}          & \textbf{3.3146}           & \textbf{3.3525}          & \textbf{3.0008}           & \textbf{3.0482  }        & \textbf{3.1714 }           & \textbf{3.2069}  \\\bottomrule        
\end{tabular}
\caption{Offline S2ST performance of StreamSpeech, evaluated with BLASER 2.0.}
\label{tab:blaser}
\end{table*}

Besides ASR-BLEU, we use BLASER 2.0\footnote{\url{https://facebookresearch.github.io/stopes/docs/eval/blaser}} to assess the quality of the generated speech. BLASER 2.0 leverages a multilingual multimodal encoder to directly encode the speech segments for source input, translation output, and reference into a shared embedding space. It then computes a score of the translation quality that can serve as a proxy for human evaluation. BLASER 2.0 comprises three versions: Unsupervised (score 0-1), QE (score 1-5), and Ref (score 1-5). Table \ref{tab:blaser} reports the offline S2ST performance of StreamSpeech evaluated by BLASER 2.0. StreamSpeech also has significant advantages over UnitY.

\begin{figure*}[t]
\centering
\subfigure[Case common\_voice\_fr\_17308913 in CVSS-C Fr$\rightarrow$En. Source transcription: \textit{laire préhistorique est le début de linformatique et est considéré comme compliquer}. Target translation: \textit{the prehistoric area is the beginning of computer science and is considered to be complicated}.]{
\includegraphics[width=\textwidth]{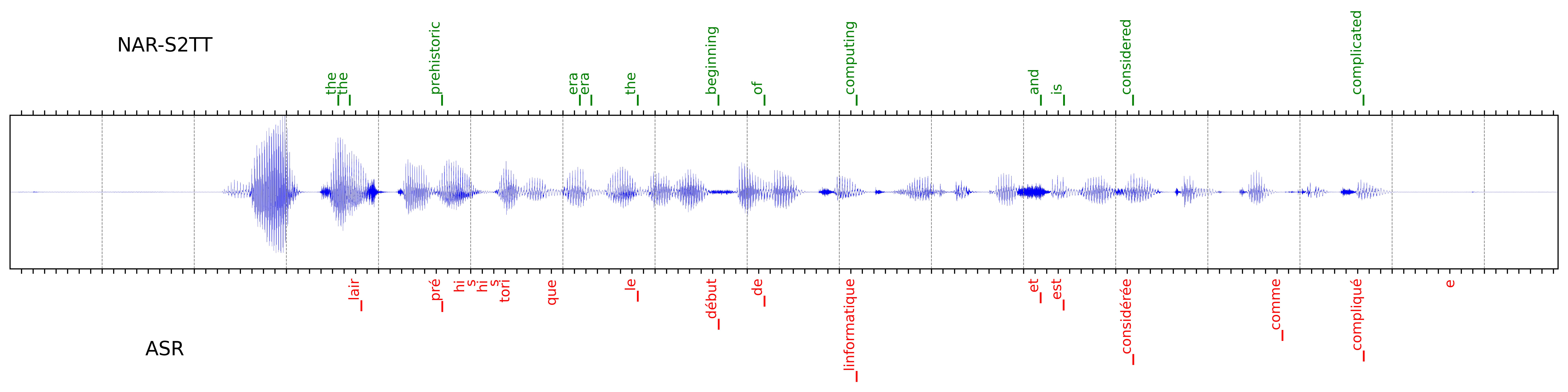}\label{fig:vis1}
}
\subfigure[Case common\_voice\_es\_18307761 in CVSS-C Es$\rightarrow$En. Source transcription: \textit{y calló  tal vez esperando una disculpa amante  pero yo preferí guardar silencio}. Target translation: \textit{and he shut up he might have been just waiting for a loving apology but i preferred to remain silent}.]{
\includegraphics[width=\textwidth]{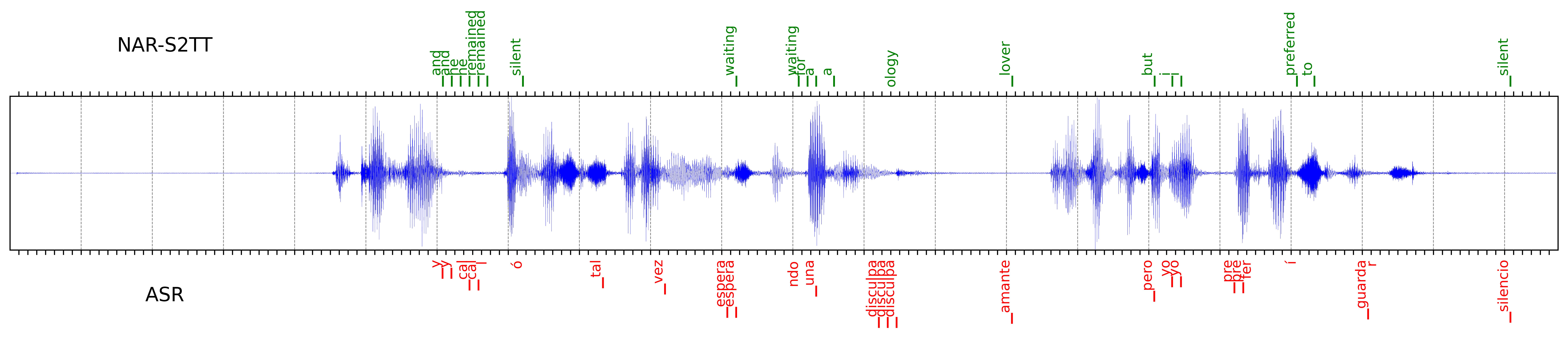}\label{fig:vis2}
}
\subfigure[Case common\_voice\_de\_17300640 in CVSS-C De$\rightarrow$En. Source transcription: \textit{dort führt eine schmale brücke über den bach}. Target translation: \textit{there a narrow bridge leads over the stream}.]{
\includegraphics[width=\textwidth]{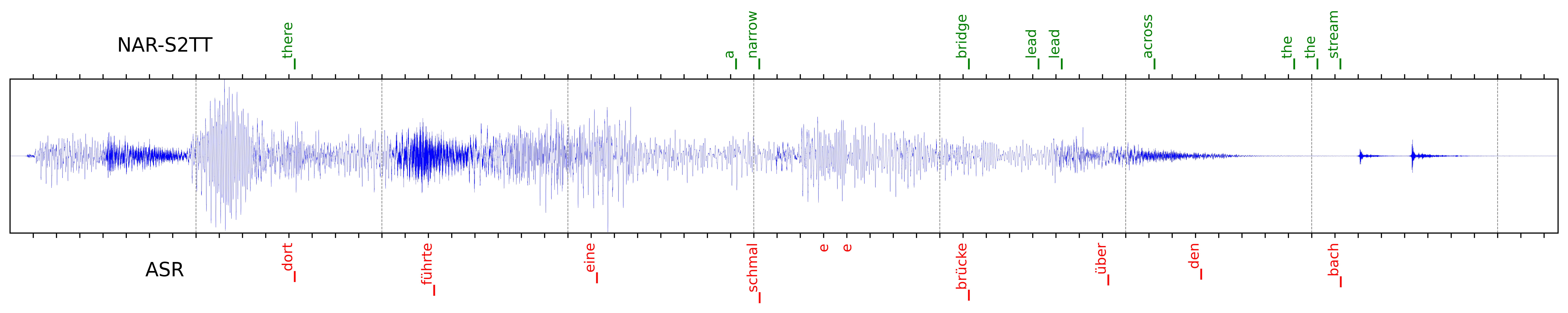}\label{fig:vis3}
}

\caption{Visualization of the alignments of source speech and source/target text within the CTC decoder for ASR and NAR-S2TT tasks. Note that the positions without label refer to generating blank token $\phi$, and we omit them for clarity. The vertical grey dashed lines represent chunks of 320$ms$.}
\label{fig:vis}
\end{figure*}

\section{Visualization of Alignments}

The policy of StreamSpeech is primarily guided by the alignment between source speech and source/target text, which is captured through the CTC decoder of the introduced ASR and NAR-S2TT tasks. We visualize the alignment captured by the CTC decoder in Figure \ref{fig:vis}.

The CTC decoder of the ASR and NAR-S2TT tasks effectively captures the alignment of speech and text and generates tokens with high accuracy, especially in terms of 1-gram accuracy. StreamSpeech starts translating upon recognizing a new source token and generates a corresponding number of target words. This ensures that the received speech before translation contains complete source tokens and provides sufficient information to generate target tokens.

Additionally, we observe that certain tokens occupying the same position in the ASR and NAR-S2TT CTC sequences correspond to the same semantic meaning. For example, `\textit{début}'$\leftrightarrow$`\textit{beginning}', `\textit{linformatique}'$\leftrightarrow$`\textit{computing}' in Fr$\rightarrow$En, `\textit{amante}'$\leftrightarrow$`\textit{lover}', `\textit{silencio}'$\leftrightarrow$`\textit{silent}' in Es$\rightarrow$En, `\textit{Dort}'$\leftrightarrow$`\textit{there}', `\textit{back}'$\leftrightarrow$`\textit{stream}' in De$\rightarrow$En. This suggests that the introduction of both source and target language CTC decoders after the encoder implicitly models cross-lingual alignments, particularly given that our introduced CTC decoder consists solely of a single fully connected layer.

\section{Case Study}

\begin{figure*}[t]
\centering
\subfigure[Streaming ASR + HMT + Real-time TTS]{
\includegraphics[width=\textwidth]{hmt_case.pdf}\label{fig:case1}
}
\subfigure[StreamSpeech]{
\includegraphics[width=\textwidth]{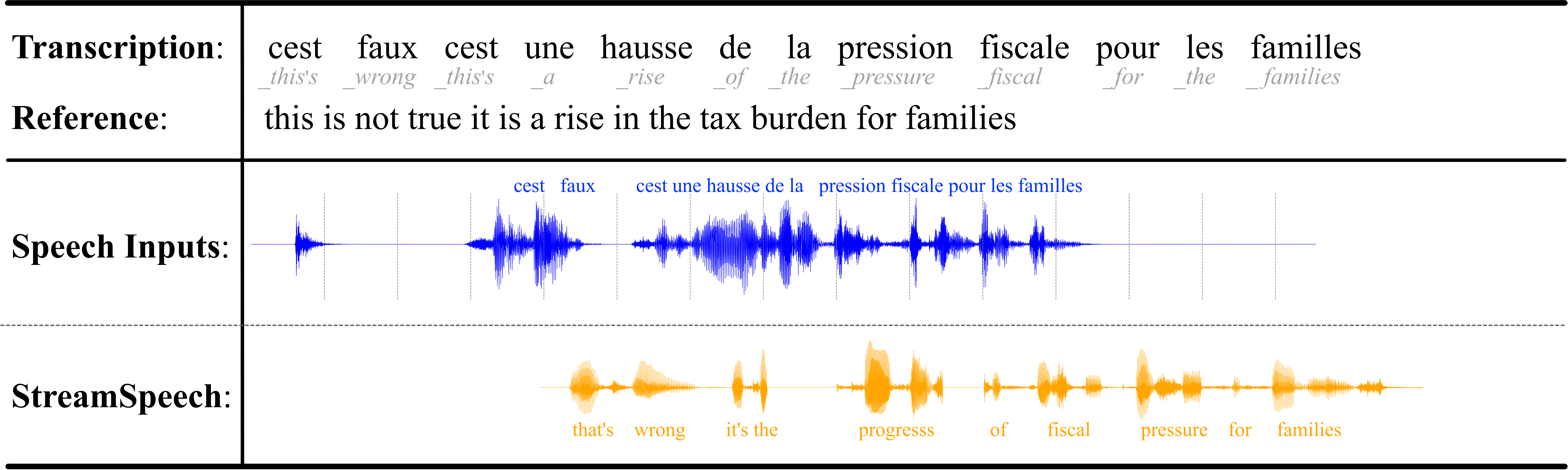}\label{fig:case2}
}

\caption{Case study of direct StreamSpeech and cascaded `ASR+HMT+TTS'. For clarity, we have marked the blue text above the source audio to represent the ground truth transcription aligned with the speech. The orange text below the target speech indicates the text transcribed by ASR-BLEU tookit.}
\label{fig:case}
\end{figure*}

In Figure \ref{fig:case}, we illustrate the Simul-S2ST process of direct StreamSpeech and the cascaded ``ASR+HMT+TTS'' system. StreamSpeech is capable of generating high-quality target speech with a delay of 2 seconds, particularly noticeable when there is prolonged silence at the beginning of the source speech. Compared to the cascaded system, direct StreamSpeech also demonstrates clear advantages. Specifically, the cascaded system requires streaming ASR to first transcribe the speech into source text, then translate the current source text into target text using state-of-the-art HMT, and finally synthesize the target speech. The cascading of multiple modules leads to error accumulation, especially as the accuracy of each module in streaming scenarios generally tends to be lower than in offline scenarios. For instance, in this case, streaming ASR may incorrectly transcribe `\textit{cest}' as `\textit{ce}', leading to subsequent HMT generating the erroneous `\textit{fake}', ultimately resulting in incorrect speech. Therefore, for more challenging simultaneous scenarios, direct models hold an advantage over cascaded systems.

You can hear more cases of StreamSpeech on offline or simultaneous speech-to-speech translation at our project page (\url{https://ictnlp.github.io/StreamSpeech-site/}).

\section{Configuration and Training Details}
\label{app:config}

Table \ref{tab:config} reports the configurations of StreamSpeech and baselines. For the offline scenario, we set the chunk size of StreamSpeech to infinity and do not involve simultaneous policy, while keeping all other settings identical to the simultaneous scenario. For the simultaneous scenario, to evaluate Simul-S2ST under different latencies, we employ multi-chunk training to train a single StreamSpeech model and utilize this model to perform Simul-S2ST under various latency. All models are trained on 4 NVIDIA RTX 3090 GPUs.

\begin{table*}[ht]
\centering\small
\begin{tabular}{L{3cm}C{1cm}ccccc} \toprule
\textbf{Hyperparameters}      & \textbf{S2UT}                                          & \textbf{Translatotron}                                     & \textbf{Translatotron2}                                    & \textbf{DASpeech}                                          & \textbf{UnitY}                                        & \textbf{StreamSpeech}                                 \\ \midrule
\multicolumn{7}{c}{\textit{Speech Encoder}}                                                                                                                                                                                                                                                                                                                                                   \\\midrule
conv\_kernel\_sizes           & (5, 5)                                                 & (5, 5)                                                     & (5, 5)                                                     & (5, 5)                                                     & (5, 5)                                                & (5, 5)                                                \\
encoder\_type                 & Conformer                                              & Conformer                                                  & Conformer                                                  & Conformer                                                  & Conformer                                             & Conformer                                             \\
encoder\_layers               & 12                                                     & 12                                                         & 12                                                         & 12                                                         & 12                                                    & 12                                                    \\
encoder\_embed\_dim           & 256                                                    & 256                                                        & 256                                                        & 256                                                        & 256                                                   & 256                                                   \\
encoder\_ffn\_embed\_dim      & 2048                                                   & 2048                                                       & 2048                                                       & 2048                                                       & 2048                                                  & 2048                                                  \\
encoder\_attention\_heads     & 4                                                      & 4                                                          & 4                                                          & 4                                                          & 4                                                     & 4                                                     \\
encoder\_pos\_enc\_type       & relative                                               & relative                                                   & relative                                                   & relative                                                   & relative                                              & relative                                              \\
depthwise\_conv\_kernel\_size & 31                                                     & 31                                                         & 31                                                         & 31                                                         & 31                                                    & 31                                                    \\
streaming                     & ×                                                      & ×                                                          & ×                                                          & ×                                                          & ×                                                     &    \checkmark                                                 \\\midrule
\multicolumn{7}{c}{\textit{Text decoder}}                                                                                                                                                                                                                                                                                                                                                              \\\midrule
decoder\_type                 & Transformer                                            & Transformer                                                & Transformer                                                & Transformer                                                & Transformer                                           & Transformer                                           \\
decoder\_layers               & 4                                                      & 4                                                          & 4                                                          & 4                                                          & 4                                                     & 4                                                     \\
decoder\_embed\_dim           & 512                                                    & 512                                                        & 512                                                        & 512                                                        & 512                                                   & 512                                                   \\
decoder\_ffn\_embed\_dim      & 2048                                                   & 2048                                                       & 2048                                                       & 2048                                                       & 2048                                                  & 2048                                                  \\
decoder\_attention\_heads     & 8                                                      & 8                                                          & 8                                                          & 8                                                          & 8                                                     & 8                                                     \\ \midrule
\multicolumn{7}{c}{\textit{Text-to-Speech Encoder}}                                                                                                                                                                                                                                                                                                                                           \\ \midrule
encoder\_type                 & -                                                      & -                                                          & Transformer                                                & -                                                          & Transformer                                           & Transformer                                           \\
encoder\_layers               & -                                                      & -                                                          & 2                                                          & -                                                          & 2                                                     & 2                                                     \\
encoder\_embed\_dim           & -                                                      & -                                                          & 512                                                        & -                                                          & 512                                                   & 512                                                   \\
encoder\_ffn\_embed\_dim      & -                                                      & -                                                          & 2048                                                       & -                                                          & 2048                                                  & 2048                                                  \\
encoder\_attention\_heads     & -                                                      & -                                                          & 8                                                          & -                                                          & 8                                                     & 8                                                     \\\midrule
\multicolumn{7}{c}{\textit{Acoustic Decoder}}                                                                                                                                                                                                                                                                                                                                                 \\\midrule
output\_type                  & \begin{tabular}[c]{@{}c@{}}unit\\ (1000)\end{tabular} & \begin{tabular}[c]{@{}c@{}}mel-\\ spectrogram\end{tabular} & \begin{tabular}[c]{@{}c@{}}mel-\\ spectrogram\end{tabular} & \begin{tabular}[c]{@{}c@{}}mel-\\ spectrogram\end{tabular} & \begin{tabular}[c]{@{}c@{}}unit\\ (1000)\end{tabular} & \begin{tabular}[c]{@{}c@{}}unit\\ (1000)\end{tabular} \\
decoder\_layers               & 6                                                      & 6                                                          & 6                                                          & 6                                                          & 2                                                     & 2                                                     \\
decoder\_embed\_dim           & 512                                                    & 512                                                        & 512                                                        & 512                                                        & 512                                                   & 512                                                   \\
decoder\_ffn\_embed\_dim      & 2048                                                   & 2048                                                       & 2048                                                       & 2048                                                       & 2048                                                  & 2048                                                  \\
decoder\_attention\_heads     & 8                                                      & 8                                                          & 8                                                          & 4                                                          & 8                                                     & 8                                                     \\\midrule
\multicolumn{7}{c}{\textit{Training}}                                                                                                                                                                                                                                                                                                                                                         \\\midrule
lr                            & 1e-3                                                   & 1e-3                                                       & 1e-3                                                       & 1e-3                                                       & 1e-3                                                  & 1e-3                                                  \\
lr\_scheduler                 & inverse\_sqrt                                          & inverse\_sqrt                                              & inverse\_sqrt                                              & inverse\_sqrt                                              & inverse\_sqrt                                         & inverse\_sqrt                                         \\
warmup\_updates               & 4000                                                   & 4000                                                       & 4000                                                       & 4000                                                       & 4000                                                  & 4000                                                  \\
warmup\_init\_lr              & 1e-7                                                   & 1e-7                                                       & 1e-7                                                       & 1e-7                                                       & 1e-7                                                  & 1e-7                                                  \\
optimizer                     & Adam                                                   & Adam                                                       & Adam                                                       & Adam                                                       & Adam                                                  & Adam                                                  \\
dropout                       & 0.1                                                    & 0.1                                                        & 0.1                                                        & 0.1                                                        & 0.1                                                   & 0.1                                                   \\
weight\_decay                 & 0.0                                                    & 0.0                                                        & 0.0                                                        & 0.0                                                        & 0.0                                                   & 0.0                                                   \\
clip\_norm                    & 1.0                                                    & 1.0                                                        & 1.0                                                        & 1.0                                                        & 1.0                                                   & 1.0                                                   \\
max\_tokens                   & 160k                                                   & 160k                                                       & 160k                                                       & 160k                                                       & 160k                                                  & 160k                                                  \\
s2st\_loss\_weight            & 1.0                                                    & 1.0                                                        & 1.0                                                        & 5.0                                                        & 1.0                                                   & 1.0                                                   \\
s2tt\_loss\_weight            & 8.0                                                    & 0.1                                                        & 0.1                                                        & 1.0                                                        & 8.0                                                   & 8.0                                                   \\
nar\_s2tt\_loss\_weight       & -                                                      & -                                                          & -                                                          & -                                                          & -                                                     & 4.0                                                   \\
asr\_loss\_weight             & -                                                      & -                                                          & -                                                          & -                                                          & -                                                     & 4.0        \\ \bottomrule                                          
\end{tabular}
\caption{Configuration of StreamSpeech and baselines.}
\label{tab:config}
\end{table*}

\section{Latency Metrics}
\label{app:metric}
To more comprehensively evaluate the latency of simultaneous speech-to-speech translation, we employ a variety of latency metrics, which are mainly divided into three categories: latency, computation-aware latency and streaming degree. 
To compute the latency metrics, we record the moment when the $i^{th}$ frame of the target speech is generated as $t_{i}$ (the starting point of the source speech considered as moment 0), and $X$ and $S$ are the source speech and target speech, respectively. Note that all metrics are automatically calculated via \textrm{SimulEval} toolkit.

Latency evaluates the duration that outputs lag behind the inputs, including:

\textbf{Average Lagging (AL)} \citep{ma-etal-2019-stacl} evaluates the average speech duration that target outputs lag behind the source inputs. AL is calculated as:
\begin{gather}
\begin{aligned}
    \mathrm{AL}=&\; \frac{1}{\tau }\sum_{i=1}^{\tau}t_{i}-\frac{i-1}{\left | S \right |/\left | X \right |},\\
    \mathrm{where} \;\;\tau = &\; \underset{i}{\mathrm{argmin}}\left ( t_{i}= \left | X \right |\right ).
\end{aligned}
\end{gather}

\textbf{Average Proportion (AP)} \citep{Cho2016} evaluates the proportion between the generating moment and the total duration of source speech, calculated as:
\begin{gather}
    \mathrm{AP}=\frac{1}{\left | X \right | \left | S \right |}\sum_{i=1}^{\left | S \right |} t_{i}.
\end{gather}

\textbf{Differentiable Average Lagging (DAL)} \citep{Arivazhagan2019} is a differentiable version of average lagging, calculated as:
\begin{gather}
\begin{aligned}
\mathrm{DAL}=&\; \frac{1}{\left | S \right | }\sum\limits_{i=1}^{\left | S \right |}t^{'}_{i}-\frac{i-1}{\left | S \right |/\left | X \right |},\\
\mathrm{where} \;\;t^{'}_{i}= &\; \left\{\begin{matrix}
t_{i} & i=1\\ 
 \mathrm{max}\left (t_{i},t^{'}_{i-1}+ \frac{\left | X \right |}{\left | S \right |} \right )& i>1
\end{matrix}\right..
\end{aligned}
\end{gather}

\textbf{StartOffset} measures the waiting time before outputting the first frame of target speech, calculated as:
\begin{gather}
    \mathrm{StartOffset}=t_{1}
\end{gather}

\textbf{EndOffset} measures the offset of the last frame of target speech relative to the end of source speech, calculated as:
\begin{gather}
    \mathrm{EndOffset}=t_{\left | S \right |}-\left | X \right |
\end{gather}

\textbf{Length-Adaptive Average Lagging (LAAL)} \citep{papi-etal-2022-generation} is a modified version of the average lagging that takes into account the over-generation phenomenon, calculated as:
\begin{gather}
\begin{aligned}
    \mathrm{LAAL}=&\; \frac{1}{\tau }\sum_{i=1}^{\tau}t_{i}-\frac{i-1}{\max(\left | S \right |,\left | S^{*} \right |)/\left | X \right |},\\
    \mathrm{where} \;\;\tau = &\; \underset{i}{\mathrm{argmin}}\left ( t_{i}= \left | X \right |\right ),
\end{aligned}
\end{gather}
where $S^{*}$ is generated target speech.

\textbf{Average Token Delay (ATD)} \citep{kano2023average}  is the average delay of output sub-segments against their corresponding input sub-segments, calculated as:
\begin{gather}
    \mathrm{ATD}= \frac{1}{\left | S \right | }\sum_{i=1}^{\left | S \right |}t_{i}-\xi_{seg_{t_{i}}} ,
\end{gather}
where $\xi_{seg_{t_{i}}}$ is the moment of corresponding input sub-segments of the ${i}^{th}$ output.

Computation-aware latency considers the actual inference time of the model when computing the aforementioned latency, including: $\mathrm{AL\_CA}$, $\mathrm{AP\_CA}$, $\mathrm{DAL\_CA}$, $\mathrm{StartOffset\_CA}$, $\mathrm{EndOffset\_CA}$, $\mathrm{LAAL\_CA}$ and $\mathrm{ATD\_CA}$.

Besides, we also evaluate the streaming degree of the generated speech. The more segments of output speech generated with shorter durations for each segment, the closer the generation is to being considered streaming. The metrics include:

\textbf{Number of Chunks (NumChunks)} evaluates the number of segments when generating the target speech.

\textbf{Discontinuity} evaluates the duration of silence produced in the generated speech while waiting for the source speech. This includes the total duration of all silences (Sum), the average duration of each silence (Ave), and the number of silence segments (Num). It's important to note that Discontinuity is not equivalent to NumChunks. When the model finishes generating a target speech segment (i.e., a chunk), if the incoming source speech is sufficient for the model to begin translation at that moment immediately, the model will not produce discontinuity.

\textbf{Real-time Factor (RTF)} \citep{fugen2007simultaneous} describes the ratio between the duration of outputs and inputs.

For more detailed implementations of latency computation, please refer to SimulEval toolkit\footnote{\url{https://github.com/facebookresearch/SimulEval/blob/main/simuleval/evaluator/scorers/latency_scorer.py}}.

\section{Numerical Results}
\label{sec:numerical}

Tables \ref{tab:numerical_fren}, \ref{tab:numerical_esen} and \ref{tab:numerical_deen} report the numerical results of StreamSpeech, including more comprehensive quality and latency metrics. 

\textbf{ASR-BLEU (with silence)}\quad In particular, we additionally calculate the ASR-BLEU considering silence for quality evaluation, denoted as \emph{ASR-BLEU (with silence)}. Specifically, StreamSpeech remains silent while waiting for the source speech after generating the current speech outputs. For instance, if StreamSpeech generates the current speech of 220$ms$, it will continue to wait for the streaming speech inputs of 320ms (corresponding to the chunk size). During the 100ms interval between 220ms and the next 320ms chunk, StreamSpeech remains silent. ASR-BLEU (with silence) is calculated directly on these speech outputs that include silence. Note that the ASR model used in ASR-BLEU was trained on standard continuous speech, which causes a mismatch when evaluating speech with silence (may lead to some recognition errors). Nevertheless, we report this metric to provide a more comprehensive evaluation of StreamSpeech.

\begin{table*}[t]
\centering\small
\begin{tabular}{L{1cm}cccC{1.7cm}C{1.7cm}C{1.7cm}cc} \toprule
\multicolumn{9}{c}{\textbf{CVSS-C Fr$\rightarrow$En}}                                                                                                                                                                                                                                                                                                                                                                                                          \\ \toprule
\multirow{2}{*}{$C\!\!\times \!\! 40ms$} & \multirow{2}{*}{\textbf{ASR-BLEU}} & \multicolumn{7}{c}{\textbf{Latency}}                                                                                                                                                                                                                                                                                                                                              \\ \cmidrule(lr){3-9}
                        &                                     & AL                                                                    & AP                   & DAL                                                                          & StartOffset                                                                  & EndOffset                                                                    & LAAL              & ATD               \\ \midrule
320$\;ms$                     & 22.89                               & 1269.84                                                               & 0.52                 & 1702.30                                                                      & 1667.11                                                                      & 699.22                                                                       & 1358.15           & 2290.69           \\
640$\;ms$                     & 24.41                               & 2326.17                                                               & 0.40                 & 1946.50                                                                      & 1888.58                                                                      & 1030.48                                                                      & 2332.34           & 2669.34           \\
960$\;ms$                     & 25.00                               & 2803.13                                                               & 0.35                 & 2124.49                                                                      & 2076.37                                                                      & 1107.48                                                                      & 2806.27           & 2862.42           \\
1280$\;ms$                    & 25.20                               & 3146.27                                                               & 0.31                 & 2309.87                                                                      & 2231.73                                                                      & 1211.81                                                                      & 3148.17           & 3079.90           \\
1600$\;ms$                    & 25.30                               & 3287.21                                                               & 0.29                 & 2352.54                                                                      & 2215.25                                                                      & 1321.50                                                                      & 3288.82           & 3240.42           \\
1920$\;ms$                    & 25.50                               & 3450.24                                                               & 0.27                 & 2477.95                                                                      & 2296.93                                                                      & 1436.70                                                                      & 3451.45           & 3381.28           \\
2240$\;ms$                    & 25.50                               & 3629.71                                                               & 0.25                 & 2666.28                                                                      & 2471.04                                                                      & 1545.23                                                                      & 3630.60           & 3520.22           \\
2560$\;ms$                    & 25.68                               & 3812.13                                                               & 0.24                 & 2891.02                                                                      & 2695.36                                                                      & 1639.45                                                                      & 3812.69           & 3651.32           \\
2880$\;ms$                    & 25.60                               & 3992.35                                                               & 0.22                 & 3131.35                                                                      & 2957.32                                                                      & 1719.35                                                                      & 3992.80           & 3776.42           \\
3200$\;ms$                    & 25.75                               & 4157.28                                                               & 0.22                 & 3370.39                                                                      & 3228.57                                                                      & 1800.14                                                                      & 4157.49           & 3908.66           \\
4800$\;ms$                    & 26.14                               & 4873.08                                                               & 0.18                 & 4505.76                                                                      & 4490.42                                                                      & 2250.83                                                                      & 4873.08           & 4640.12           \\
10000$\;ms$                  & 26.20                               & 5683.92                                                               & 0.13                 & 5683.92                                                                      & 5683.92                                                                      & 3096.54                                                                      & 5683.92           & 5672.35           \\ \midrule
\multirow{2}{*}{$C\!\!\times \!\! 40ms$} & \multirow{2}{*}{\textbf{ASR-BLEU}} & \multicolumn{7}{c}{\textbf{Computation-Aware Latency}}                                                                                                                                                                                                                                                                                                                            \\\cmidrule(lr){3-9}
                        &                                     & AL\_CA                                                                & AP\_CA               & DAL\_CA                                                                      & StartOffset\_CA                                                              & EndOffset\_CA                                                                & LAAL\_CA          & ATD\_CA           \\ \midrule
320$\;ms$                     & 22.89                               & 2195.04                                                               & 0.68                 & 2333.37                                                                      & 2052.33                                                                      & 699.22                                                                       & 2269.02           & 2840.15           \\
640$\;ms$                     & 24.41                               & 2908.52                                                               & 0.47                 & 2369.64                                                                      & 2164.57                                                                      & 1030.48                                                                      & 2913.75           & 2958.70           \\
960$\;ms$                     & 25.00                               & 3331.47                                                               & 0.41                 & 2548.80                                                                      & 2363.50                                                                      & 1107.48                                                                      & 3334.27           & 3133.06           \\
1280$\;ms$                    & 25.20                               & 3576.95                                                               & 0.35                 & 2680.58                                                                      & 2469.84                                                                      & 1211.81                                                                      & 3578.67           & 3292.64           \\
1600$\;ms$                    & 25.30                               & 3694.33                                                               & 0.33                 & 2756.06                                                                      & 2451.59                                                                      & 1321.50                                                                      & 3695.72           & 3449.34           \\
1920$\;ms$                    & 25.50                               & 3765.89                                                               & 0.29                 & 2777.31                                                                      & 2465.22                                                                      & 1436.70                                                                      & 3766.86           & 3540.84           \\
2240$\;ms$                    & 25.50                               & 3995.11                                                               & 0.28                 & 3029.87                                                                      & 2677.67                                                                      & 1545.23                                                                      & 3995.87           & 3725.77           \\
2560$\;ms$                    & 25.68                               & 4167.48                                                               & 0.26                 & 3229.14                                                                      & 2920.50                                                                      & 1639.45                                                                      & 4168.00           & 3837.85           \\
2880$\;ms$                    & 25.60                               & 4323.54                                                               & 0.24                 & 3430.48                                                                      & 3170.37                                                                      & 1719.35                                                                      & 4323.93           & 3951.69           \\
3200$\;ms$                    & 25.75                               & 4502.11                                                               & 0.23                 & 3671.07                                                                      & 3406.31                                                                      & 1800.14                                                                      & 4502.32           & 4105.75           \\
4800$\;ms$                    & 26.14                               & 5189.20                                                               & 0.19                 & 4752.24                                                                      & 4723.36                                                                      & 2250.83                                                                      & 5189.20           & 4815.85           \\
10000$\;ms$                  & 26.20                               & 5946.97                                                               & 0.13                 & 5946.97                                                                      & 5946.97                                                                      & 3096.54                                                                      & 5946.97           & 5816.81           \\ \midrule
\multirow{3}{*}{$C\!\!\times \!\! 40ms$} & \multirow{3}{*}{\textbf{ASR-BLEU}} & \multicolumn{7}{c}{\textbf{Streaming Degree}}                                                                                                                                                                                                                                                                                                                                     \\\cmidrule(lr){3-9}
                        &                                     & \multirow{2}{*}{\begin{tabular}[c]{@{}c@{}}Num\\ Chunks\end{tabular}} & \multirow{2}{*}{RTF} & \multirow{2}{*}{\begin{tabular}[c]{@{}c@{}}Discontinuity\\ Sum\end{tabular}} & \multirow{2}{*}{\begin{tabular}[c]{@{}c@{}}Discontinuity\\ Ave\end{tabular}} & \multirow{2}{*}{\begin{tabular}[c]{@{}c@{}}Discontinuity\\ Num\end{tabular}} & \multirow{2}{*}{} & \multirow{2}{*}{} \\ 
                        &                                     &                                                                       &                      &                                                                              &                                                                              &                                                                              &                   &                   \\ \midrule
320$\;ms$                     & 22.89                               & 7.85                                                                  & 1.15                 & 1695.99                                                                      & 385.77                                                                       & 4.42                                                                         &                   &                   \\
640$\;ms$                     & 24.41                               & 5.43                                                                  & 1.20                 & 1745.35                                                                      & 549.87                                                                       & 3.28                                                                         &                   &                   \\
960$\;ms$                     & 25.00                               & 4.46                                                                  & 1.22                 & 1630.69                                                                      & 585.05                                                                       & 2.72                                                                         &                   &                   \\
1280$\;ms$                    & 25.20                               & 3.83                                                                  & 1.24                 & 1583.53                                                                      & 672.25                                                                       & 2.23                                                                         &                   &                   \\
1600$\;ms$                    & 25.30                               & 3.48                                                                  & 1.26                 & 1705.61                                                                      & 843.01                                                                       & 1.93                                                                         &                   &                   \\
1920$\;ms$                    & 25.50                               & 3.15                                                                  & 1.29                 & 1736.51                                                                      & 997.20                                                                       & 1.64                                                                         &                   &                   \\
2240$\;ms$                    & 25.50                               & 2.86                                                                  & 1.30                 & 1668.21                                                                      & 1101.78                                                                      & 1.37                                                                         &                   &                   \\
2560$\;ms$                    & 25.68                               & 2.62                                                                  & 1.32                 & 1543.38                                                                      & 1144.47                                                                    & 1.16                                                                         &                   &                   \\
2880$\;ms$                    & 25.60                               & 2.40                                                                  & 1.34                 & 1361.69                                                                      & 1103.53                                                                      & 0.97                                                                         &                   &                   \\
3200$\;ms$                    & 25.75                               & 2.23                                                                  & 1.35                 & 1166.73                                                                      & 1009.29                                                                      & 0.80                                                                         &                   &                   \\
4800$\;ms$                    & 26.14                               & 1.69                                                                  & 1.44                 & 356.97                                                                       & 354.08                                                                       & 0.25                                                                         &                   &                   \\
10000$\;ms$                  & 26.20                               & 1.00                                                                  & 1.56                 & 0.00                                                                         & 0.00                                                                         & 0.00                                                                         &                   &          \\\midrule
\multirow{3}{*}{$C\!\!\times \!\! 40ms$} & \multirow{3}{*}{\textbf{ASR-BLEU}}  & \multicolumn{7}{c}{\textbf{Quality with other Metrics}}                                                                                                                                                                                                                                                                                                                                                     \\\cmidrule(lr){3-9}
                        &                            & \multicolumn{2}{c}{\multirow{2}{*}{\begin{tabular}[c]{@{}c@{}}ASR-BLEU\\ (with silence)\end{tabular}}} & \multirow{2}{*}{\begin{tabular}[c]{@{}c@{}}BLASER 2.0\\ (Unsupervised)\end{tabular}} & \multirow{2}{*}{\begin{tabular}[c]{@{}c@{}}BLASER 2.0\\ (QE)\end{tabular}}   & \multirow{2}{*}{\begin{tabular}[c]{@{}c@{}}BLASER 2.0\\ (Ref)\end{tabular}}  &                   &                   \\
                        &                            & \multicolumn{2}{c}{}                                                                                 &                                                                                      &                                                                              &                                                                              &                   &                   \\ \midrule
320$\;ms$                     & 22.89                      & \multicolumn{2}{c}{17.88}                                                                            & 0.4428                                                                               & 3.1170                                                                       & 3.0519                                                                       &                   &                   \\
640$\;ms$                     & 24.41                      & \multicolumn{2}{c}{19.65}                                                                            & 0.4439                                                                               & 3.1322                                                                       & 3.0965                                                                       &                   &                   \\
960$\;ms$                     & 25.00                      & \multicolumn{2}{c}{20.20}                                                                            & 0.4446                                                                               & 3.1373                                                                       & 3.1067                                                                       &                   &                   \\
1280$\;ms$                    & 25.20                      & \multicolumn{2}{c}{21.14}                                                                            & 0.4448                                                                               & 3.1391                                                                       & 3.1109                                                                       &                   &                   \\
1600$\;ms$                    & 25.30                      & \multicolumn{2}{c}{21.07}                                                                            & 0.4450                                                                               & 3.1420                                                                       & 3.1164                                                                       &                   &                   \\
1920$\;ms$                    & 25.50                      & \multicolumn{2}{c}{21.61}                                                                            & 0.4451                                                                               & 3.1429                                                                       & 3.1195                                                                       &                   &                   \\
2240$\;ms$                    & 25.50                      & \multicolumn{2}{c}{21.97}                                                                            & 0.4451                                                                               & 3.1443                                                                       & 3.1226                                                                       &                   &                   \\
2560$\;ms$                    & 25.68                      & \multicolumn{2}{c}{22.76}                                                                            & 0.4456                                                                               & 3.1464                                                                       & 3.1268                                                                       &                   &                   \\
2880$\;ms$                    & 25.60                      & \multicolumn{2}{c}{23.26}                                                                            & 0.4456                                                                               & 3.1484                                                                       & 3.1294                                                                       &                   &                   \\
3200$\;ms$                    & 25.75                      & \multicolumn{2}{c}{23.91}                                                                            & 0.4456                                                                               & 3.1475                                                                       & 3.1294                                                                       &                   &                   \\
4800$\;ms$                    & 26.14                      & \multicolumn{2}{c}{25.66}                                                                            & 0.4462                                                                               & 3.1530                                                                       & 3.1397                                                                       &                   &                   \\
10000$\;ms$                  & 26.20                      & \multicolumn{2}{c}{26.20}                                                                            & 0.4465                                                                               & 3.1569                                                                       & 3.1486                                                                       &                   & \\\bottomrule        
\end{tabular}
\caption{Numerical results of StreamSpeech on CVSS-C Fr$\rightarrow$En.}
\label{tab:numerical_fren}
\end{table*}

\begin{table*}[t]
\centering\small
\begin{tabular}{L{1cm}cccC{1.7cm}C{1.7cm}C{1.7cm}cc} \toprule
\multicolumn{9}{c}{\textbf{CVSS-C Es$\rightarrow$En}}                                                                                                                                                                                                                                                                                                                                                                                                                 \\ \toprule
\multirow{2}{*}{$C\!\!\times \!\! 40ms$} & \multirow{2}{*}{\textbf{ASR-BLEU}} & \multicolumn{7}{c}{\textbf{Latency}}                                                                                                                                                                                                                                                                                                                                              \\\cmidrule(lr){3-9}
                                 &                                    & AL                                                                    & AP                   & DAL                                                                          & StartOffset                                                                  & EndOffset                                                                    & LAAL              & ATD               \\ \midrule
320$\;ms$                              & 20.06                              & 1522.05                                                               & 0.52                 & 1899.15                                                                      & 1829.94                                                                      & 811.60                                                                       & 1611.94           & 2647.52           \\
640$\;ms$                              & 21.68                              & 2514.69                                                               & 0.40                 & 2129.15                                                                      & 2050.91                                                                      & 1082.63                                                                      & 2522.64           & 3000.61           \\
960$\;ms$                              & 22.36                              & 2999.86                                                               & 0.35                 & 2274.76                                                                      & 2207.81                                                                      & 1138.16                                                                      & 3002.95           & 3189.89           \\
1280$\;ms$                             & 22.76                              & 3410.10                                                               & 0.31                 & 2510.28                                                                      & 2438.99                                                                      & 1218.14                                                                      & 3411.50           & 3400.50           \\
1600$\;ms$                             & 22.94                              & 3577.51                                                               & 0.29                 & 2566.79                                                                      & 2433.17                                                                      & 1310.75                                                                      & 3578.60           & 3571.73           \\
1920$\;ms$                             & 23.19                              & 3708.04                                                               & 0.27                 & 2632.80                                                                      & 2442.63                                                                      & 1423.14                                                                      & 3709.03           & 3716.76           \\
2240$\;ms$                             & 23.26                              & 3870.11                                                               & 0.25                 & 2785.82                                                                      & 2564.04                                                                      & 1516.54                                                                      & 3870.45           & 3846.65           \\
2560$\;ms$                             & 23.46                              & 4050.40                                                               & 0.23                 & 2992.91                                                                      & 2766.25                                                                      & 1616.86                                                                      & 4050.55           & 3982.72           \\
2880$\;ms$                             & 23.51                              & 4236.50                                                               & 0.22                 & 3232.40                                                                      & 3019.76                                                                      & 1694.92                                                                      & 4236.58           & 4108.61           \\
3200$\;ms$                             & 23.58                              & 4408.96                                                               & 0.21                 & 3476.29                                                                      & 3289.87                                                                      & 1766.69                                                                      & 4409.03           & 4227.52           \\
4800$\;ms$                             & 23.97                              & 5161.83                                                               & 0.18                 & 4677.54                                                                      & 4654.39                                                                      & 2131.72                                                                      & 5161.83           & 4909.06           \\
10000$\;ms$                            & 24.22                              & 6185.38                                                               & 0.12                 & 6185.38                                                                      & 6185.38                                                                      & 3118.92                                                                      & 6185.38           & 6171.87           \\ \midrule
\multirow{2}{*}{$C\!\!\times \!\! 40ms$} & \multirow{2}{*}{\textbf{ASR-BLEU}} & \multicolumn{7}{c}{\textbf{Computation-Aware Latency}}                                                                                                                                                                                                                                                                                                                            \\\cmidrule(lr){3-9}
                                 &                                    & AL\_CA                                                                & AP\_CA               & DAL\_CA                                                                      & StartOffset\_CA                                                              & EndOffset\_CA                                                                & LAAL\_CA          & ATD\_CA           \\\midrule
320$\;ms$                              & 20.06                              & 2395.05                                                               & 0.66                 & 2474.80                                                                      & 2177.62                                                                      & 811.60                                                                       & 2471.77           & 3041.25           \\
640$\;ms$                              & 21.68                              & 3224.90                                                               & 0.49                 & 2751.52                                                                      & 2425.16                                                                      & 1082.63                                                                      & 3231.63           & 3428.89           \\
960$\;ms$                              & 22.36                              & 3462.63                                                               & 0.40                 & 2625.00                                                                      & 2414.63                                                                      & 1138.16                                                                      & 3465.45           & 3410.61           \\
1280$\;ms$                             & 22.76                              & 3826.98                                                               & 0.35                 & 2844.11                                                                      & 2653.91                                                                      & 1218.14                                                                      & 3828.28           & 3590.27           \\
1600$\;ms$                             & 22.94                              & 3974.31                                                               & 0.32                 & 2945.91                                                                      & 2637.41                                                                      & 1310.75                                                                      & 3975.25           & 3782.11           \\
1920$\;ms$                             & 23.19                              & 4048.74                                                               & 0.29                 & 2964.03                                                                      & 2622.29                                                                      & 1423.14                                                                      & 4049.54           & 3895.00           \\
2240$\;ms$                             & 23.26                              & 4211.71                                                               & 0.27                 & 3124.19                                                                      & 2746.27                                                                      & 1516.54                                                                      & 4212.04           & 4035.37           \\
2560$\;ms$                             & 23.46                              & 4348.39                                                               & 0.25                 & 3282.05                                                                      & 2912.61                                                                      & 1616.86                                                                      & 4348.52           & 4151.34           \\
2880$\;ms$                             & 23.51                              & 4556.30                                                               & 0.24                 & 3521.85                                                                      & 3176.94                                                                      & 1694.92                                                                      & 4556.37           & 4291.52           \\
3200$\;ms$                             & 23.58                              & 4725.99                                                               & 0.23                 & 3765.53                                                                      & 3468.13                                                                      & 1766.69                                                                      & 4726.06           & 4412.92           \\
4800$\;ms$                             & 23.97                              & 5487.24                                                               & 0.19                 & 4921.40                                                                      & 4888.64                                                                      & 2131.72                                                                      & 5487.24           & 5086.99           \\
10000$\;ms$                            & 24.22                              & 6472.57                                                               & 0.12                 & 6472.57                                                                      & 6472.57                                                                      & 3118.92                                                                      & 6472.57           & 6329.43           \\\midrule
\multirow{3}{*}{$C\!\!\times \!\! 40ms$} & \multirow{3}{*}{\textbf{ASR-BLEU}} & \multicolumn{7}{c}{\textbf{Streaming Degree}}                                                                                                                                                                                                                                                                                                                                     \\\cmidrule(lr){3-9}
                                 &                                    & \multirow{2}{*}{\begin{tabular}[c]{@{}c@{}}Num\\ Chunks\end{tabular}} & \multirow{2}{*}{RTF} & \multirow{2}{*}{\begin{tabular}[c]{@{}c@{}}Discontinuity\\ Sum\end{tabular}} & \multirow{2}{*}{\begin{tabular}[c]{@{}c@{}}Discontinuity\\ Ave\end{tabular}} & \multirow{2}{*}{\begin{tabular}[c]{@{}c@{}}Discontinuity\\ Num\end{tabular}} & \multirow{2}{*}{} & \multirow{2}{*}{} \\
                                 &                                    &                                                                       &                      &                                                                              &                                                                              &                                                                              &                   &                   \\\midrule
320$\;ms$                              & 20.06                              & 8.02                                                                  & 1.15                 & 2141.84                                                                      & 455.29                                                                       & 5.01                                                                         &                   &                   \\
640$\;ms$                              & 21.68                              & 5.74                                                                  & 1.19                 & 2119.97                                                                      & 600.15                                                                       & 3.81                                                                         &                   &                   \\
960$\;ms$                              & 22.36                              & 4.75                                                                  & 1.20                 & 2010.21                                                                      & 659.49                                                                       & 3.16                                                                         &                   &                   \\
1280$\;ms$                             & 22.76                              & 4.02                                                                  & 1.21                 & 1857.09                                                                      & 728.14                                                                       & 2.52                                                                         &                   &                   \\
1600$\;ms$                             & 22.94                              & 3.63                                                                  & 1.23                 & 1953.86                                                                      & 880.22                                                                       & 2.16                                                                         &                   &                   \\
1920$\;ms$                             & 23.19                              & 3.33                                                                  & 1.25                 & 2050.74                                                                      & 1068.46                                                                      & 1.88                                                                         &                   &                   \\
2240$\;ms$                             & 23.26                              & 3.05                                                                  & 1.27                 & 2024.67                                                                      & 1226.73                                                                      & 1.60                                                                         &                   &                   \\
2560$\;ms$                             & 23.46                              & 2.79                                                                  & 1.28                 & 1921.86                                                                      & 1326.79                                                                      & 1.36                                                                         &                   &                   \\
2880$\;ms$                             & 23.51                              & 2.56                                                                  & 1.29                 & 1751.37                                                                      & 1357.80                                                                      & 1.15                                                                         &                   &                   \\
3200$\;ms$                             & 23.58                              & 2.38                                                                  & 1.30                 & 1546.32                                                                      & 1298.67                                                                      & 0.99                                                                         &                   &                   \\
4800$\;ms$                             & 23.97                              & 1.81                                                                  & 1.37                 & 539.01                                                                       & 534.01                                                                       & 0.37                                                                         &                   &                   \\
10000$\;ms$                            & 24.22                              & 1.00                                                                  & 1.52                 & 0.00                                                                         & 0.00                                                                         & 0.00                                                                         &                   &        \\ \midrule
\multirow{3}{*}{$C\!\!\times \!\! 40ms$} & \multirow{3}{*}{\textbf{ASR-BLEU}}  & \multicolumn{7}{c}{\textbf{Quality with other Metrics}}                                                                                                                                                                                                                                                                                                                                                     \\ \cmidrule(lr){3-9}
                        &                            & \multicolumn{2}{c}{\multirow{2}{*}{\begin{tabular}[c]{@{}c@{}}ASR-BLEU\\ (with silence)\end{tabular}}} & \multirow{2}{*}{\begin{tabular}[c]{@{}c@{}}BLASER 2.0\\ (Unsupervised)\end{tabular}} & \multirow{2}{*}{\begin{tabular}[c]{@{}c@{}}BLASER 2.0\\ (QE)\end{tabular}}   & \multirow{2}{*}{\begin{tabular}[c]{@{}c@{}}BLASER 2.0\\ (Ref)\end{tabular}}  &                   &                   \\
                        &                            & \multicolumn{2}{c}{}                                                                                 &                                                                                      &                                                                              &                                                                              &                   &                   \\\midrule
320$\;ms$                     & 20.06                      & \multicolumn{2}{c}{14.76}                                                                            & 0.5011                                                                               & 3.2251                                                                       & 3.0946                                                                       &                   &                   \\
640$\;ms$                     & 21.68                      & \multicolumn{2}{c}{16.57}                                                                            & 0.5053                                                                               & 3.2593                                                                       & 3.1567                                                                       &                   &                   \\
960$\;ms$                     & 22.36                      & \multicolumn{2}{c}{17.49}                                                                            & 0.5072                                                                               & 3.2722                                                                       & 3.1783                                                                       &                   &                   \\
1280$\;ms$                    & 22.76                      & \multicolumn{2}{c}{18.29}                                                                            & 0.5074                                                                               & 3.2767                                                                       & 3.1856                                                                       &                   &                   \\
1600$\;ms$                    & 22.94                      & \multicolumn{2}{c}{18.64}                                                                            & 0.5083                                                                               & 3.2812                                                                       & 3.1921                                                                       &                   &                   \\
1920$\;ms$                    & 23.19                      & \multicolumn{2}{c}{18.93}                                                                            & 0.5085                                                                               & 3.2842                                                                       & 3.1994                                                                       &                   &                   \\
2240$\;ms$                    & 23.26                      & \multicolumn{2}{c}{19.29}                                                                            & 0.5089                                                                               & 3.2866                                                                       & 3.2052                                                                       &                   &                   \\
2560$\;ms$                    & 23.46                      & \multicolumn{2}{c}{20.12}                                                                            & 0.5093                                                                               & 3.2908                                                                       & 3.2099                                                                       &                   &                   \\
2880$\;ms$                    & 23.51                      & \multicolumn{2}{c}{20.80}                                                                            & 0.5099                                                                               & 3.2943                                                                       & 3.2157                                                                       &                   &                   \\
3200$\;ms$                    & 23.58                      & \multicolumn{2}{c}{21.22}                                                                            & 0.5102                                                                               & 3.2957                                                                       & 3.2176                                                                       &                   &                   \\
4800$\;ms$                    & 23.97                      & \multicolumn{2}{c}{23.29}                                                                            & 0.5108                                                                               & 3.3017                                                                       & 3.2295                                                                       &                   &                   \\
10000$\;ms$                  & 24.22                      & \multicolumn{2}{c}{24.22}                                                                            & 0.5114                                                                               & 3.3075                                                                       & 3.2397                                                                       &                   &  \\\bottomrule          
\end{tabular}
\caption{Numerical results of StreamSpeech on CVSS-C Es$\rightarrow$En.}
\label{tab:numerical_esen}
\end{table*}

\begin{table*}[t]
\centering\small
\begin{tabular}{L{1cm}cccC{1.7cm}C{1.7cm}C{1.7cm}cc} \toprule
\multicolumn{9}{c}{\textbf{CVSS-C De$\rightarrow$En}}                                                                                                                                                                                                                                                                                                                                                                                                                 \\ \toprule
\multirow{2}{*}{$C\!\!\times \!\! 40ms$} & \multirow{2}{*}{\textbf{ASR-BLEU}} & \multicolumn{7}{c}{\textbf{Latency}}                                                                                                                                                                                                                                                                                                                                              \\\cmidrule(lr){3-9}
                                 &                                    & AL                                                                    & AP                   & DAL                                                                          & StartOffset                                                                  & EndOffset                                                                    & LAAL              & ATD               \\ \midrule
320$\;ms$                              & 14.56                              & 1687.62                                                               & 0.46                 & 1815.81                                                                      & 1758.31                                                                      & 1186.14                                                                      & 1741.47           & 2736.31           \\
640$\;ms$                              & 15.83                              & 2561.63                                                               & 0.36                 & 2078.04                                                                      & 2004.57                                                                      & 1327.14                                                                      & 2566.84           & 3042.11           \\
960$\;ms$                              & 16.34                              & 2978.14                                                               & 0.32                 & 2256.35                                                                      & 2189.65                                                                      & 1361.72                                                                      & 2980.68           & 3202.18           \\
1280$\;ms$                             & 16.57                              & 3276.92                                                               & 0.29                 & 2424.27                                                                      & 2341.71                                                                      & 1426.29                                                                      & 3279.19           & 3379.61           \\
1600$\;ms$                             & 16.75                              & 3418.49                                                               & 0.28                 & 2477.75                                                                      & 2341.77                                                                      & 1506.44                                                                      & 3420.34           & 3516.78           \\
1920$\;ms$                             & 16.85                              & 3568.48                                                               & 0.26                 & 2597.41                                                                      & 2426.33                                                                      & 1587.89                                                                      & 3569.64           & 3640.85           \\
2240$\;ms$                             & 17.02                              & 3736.77                                                               & 0.24                 & 2777.56                                                                      & 2591.56                                                                      & 1668.12                                                                      & 3737.84           & 3758.86           \\
2560$\;ms$                             & 17.17                              & 3904.82                                                               & 0.23                 & 2982.85                                                                      & 2800.69                                                                      & 1733.56                                                                      & 3905.40           & 3868.65           \\
2880$\;ms$                             & 17.23                              & 4060.73                                                               & 0.22                 & 3193.25                                                                      & 3027.93                                                                      & 1802.33                                                                      & 4061.25           & 3984.06           \\
3200$\;ms$                             & 17.16                              & 4219.67                                                               & 0.21                 & 3425.01                                                                      & 3282.92                                                                      & 1862.79                                                                      & 4220.07           & 4102.51           \\
4800$\;ms$                             & 17.52                              & 4916.03                                                               & 0.18                 & 4519.45                                                                      & 4500.36                                                                      & 2205.11                                                                      & 4916.18           & 4736.56           \\
10000$\;ms$                            & 18.05                              & 5741.25                                                               & 0.12                 & 5741.25                                                                      & 5741.25                                                                      & 2968.60                                                                      & 5741.25           & 5730.22           \\ \midrule
\multirow{2}{*}{$C\!\!\times \!\! 40ms$} & \multirow{2}{*}{\textbf{ASR-BLEU}} & \multicolumn{7}{c}{\textbf{Computation-Aware Latency}}                                                                                                                                                                                                                                                                                                                            \\ \cmidrule(lr){3-9}
                                 &                                    & AL\_CA                                                                & AP\_CA               & DAL\_CA                                                                      & StartOffset\_CA                                                              & EndOffset\_CA                                                                & LAAL\_CA          & ATD\_CA           \\ \midrule
320$\;ms$                              & 14.56                              & 2558.81                                                               & 0.59                 & 2446.13                                                                      & 2135.38                                                                      & 1186.14                                                                      & 2604.99           & 3196.74           \\
640$\;ms$                              & 15.83                              & 3161.91                                                               & 0.43                 & 2524.26                                                                      & 2284.92                                                                      & 1327.14                                                                      & 3166.60           & 3338.35           \\
960$\;ms$                              & 16.34                              & 3448.73                                                               & 0.36                 & 2590.87                                                                      & 2418.56                                                                      & 1361.72                                                                      & 3450.96           & 3407.84           \\
1280$\;ms$                             & 16.57                              & 3677.11                                                               & 0.33                 & 2735.80                                                                      & 2552.65                                                                      & 1426.29                                                                      & 3679.01           & 3561.59           \\
1600$\;ms$                             & 16.75                              & 3752.73                                                               & 0.30                 & 2764.69                                                                      & 2510.66                                                                      & 1506.44                                                                      & 3754.44           & 3681.01           \\
1920$\;ms$                             & 16.85                              & 3883.72                                                               & 0.28                 & 2901.05                                                                      & 2597.63                                                                      & 1587.89                                                                      & 3884.82           & 3805.99           \\
2240$\;ms$                             & 17.02                              & 4009.48                                                               & 0.26                 & 3024.46                                                                      & 2736.61                                                                      & 1668.12                                                                      & 4010.50           & 3905.44           \\
2560$\;ms$                             & 17.17                              & 4174.97                                                               & 0.25                 & 3221.65                                                                      & 2952.58                                                                      & 1733.56                                                                      & 4175.52           & 4014.60           \\
2880$\;ms$                             & 17.23                              & 4339.98                                                               & 0.24                 & 3444.19                                                                      & 3203.36                                                                      & 1802.33                                                                      & 4340.44           & 4147.01           \\
3200$\;ms$                             & 17.16                              & 4484.03                                                               & 0.23                 & 3654.72                                                                      & 3448.12                                                                      & 1862.79                                                                      & 4484.39           & 4253.07           \\
4800$\;ms$                             & 17.52                              & 5234.16                                                               & 0.19                 & 4766.98                                                                      & 4741.92                                                                      & 2205.11                                                                      & 5234.31           & 4912.98           \\
10000$\;ms$                            & 18.05                              & 5977.26                                                               & 0.13                 & 5977.26                                                                      & 5977.26                                                                      & 2968.60                                                                      & 5977.26           & 5860.21           \\ \midrule
\multirow{3}{*}{$C\!\!\times \!\! 40ms$} & \multirow{3}{*}{\textbf{ASR-BLEU}} & \multicolumn{7}{c}{\textbf{Streaming Degree}}                                                                                                                                                                                                                                                                                                                                     \\\cmidrule(lr){3-9}
                                 &                                    & \multirow{2}{*}{\begin{tabular}[c]{@{}c@{}}Num\\ Chunks\end{tabular}} & \multirow{2}{*}{RTF} & \multirow{2}{*}{\begin{tabular}[c]{@{}c@{}}Discontinuity\\ Sum\end{tabular}} & \multirow{2}{*}{\begin{tabular}[c]{@{}c@{}}Discontinuity\\ Ave\end{tabular}} & \multirow{2}{*}{\begin{tabular}[c]{@{}c@{}}Discontinuity\\ Num\end{tabular}} & \multirow{2}{*}{} & \multirow{2}{*}{} \\
                                 &                                    &                                                                       &                      &                                                                              &                                                                              &                                                                              &                   &                   \\\midrule
320$\;ms$                              & 14.56                              & 6.85                                                                  & 1.24                 & 2246.02                                                                      & 565.17                                                                       & 4.39                                                                         &                   &                   \\
640$\;ms$                              & 15.83                              & 4.93                                                                  & 1.26                 & 2092.95                                                                      & 703.44                                                                       & 3.24                                                                         &                   &                   \\
960$\;ms$                              & 16.34                              & 4.15                                                                  & 1.27                 & 1942.50                                                                      & 734.28                                                                       & 2.71                                                                         &                   &                   \\
1280$\;ms$                             & 16.57                              & 3.64                                                                  & 1.28                 & 1851.17                                                                      & 800.32                                                                       & 2.25                                                                         &                   &                   \\
1600$\;ms$                             & 16.75                              & 3.34                                                                  & 1.30                 & 1926.99                                                                      & 948.35                                                                       & 1.96                                                                         &                   &                   \\
1920$\;ms$                             & 16.85                              & 3.06                                                                  & 1.31                 & 1922.77                                                                      & 1074.24                                                                      & 1.69                                                                         &                   &                   \\
2240$\;ms$                             & 17.02                              & 2.81                                                                  & 1.33                 & 1832.01                                                                      & 1165.29                                                                      & 1.43                                                                         &                   &                   \\
2560$\;ms$                             & 17.17                              & 2.58                                                                  & 1.34                 & 1695.21                                                                      & 1200.79                                                                      & 1.22                                                                         &                   &                   \\
2880$\;ms$                             & 17.23                              & 2.39                                                                  & 1.35                 & 1531.03                                                                      & 1182.17                                                                      & 1.04                                                                         &                   &                   \\
3200$\;ms$                             & 17.16                              & 2.23                                                                  & 1.36                 & 1339.35                                                                      & 1112.27                                                                      & 0.88                                                                         &                   &                   \\
4800$\;ms$                             & 17.52                              & 1.68                                                                  & 1.43                 & 482.26                                                                       & 478.56                                                                       & 0.32                                                                         &                   &                   \\
10000$\;ms$                            & 18.05                              & 1.00                                                                  & 1.54                 & 0.00                                                                         & 0.00                                                                         & 0.00                                                                         &                   &          \\ \midrule
\multirow{3}{*}{$C\!\!\times \!\! 40ms$} & \multirow{3}{*}{\textbf{ASR-BLEU}} & \multicolumn{7}{c}{\textbf{Quality with other Metrics}}                                                                                                                                                                                                                                                                                                                                             \\ \cmidrule(lr){3-9}
                        &                                    & \multicolumn{2}{c}{\multirow{2}{*}{\begin{tabular}[c]{@{}c@{}}ASR-BLEU\\ (with silence)\end{tabular}}} & \multirow{2}{*}{\begin{tabular}[c]{@{}c@{}}BLASER 2.0\\ (Unsupervised)\end{tabular}} & \multirow{2}{*}{\begin{tabular}[c]{@{}c@{}}BLASER 2.0\\ (QE)\end{tabular}}   & \multirow{2}{*}{\begin{tabular}[c]{@{}c@{}}BLASER 2.0\\ (Ref)\end{tabular}}  &                   &                   \\
                        &                                    & \multicolumn{2}{c}{}                                                                                   &                                                                                      &                                                                              &                                                                              &                   &                   \\ \midrule
320$\;ms$                       & 14.56                              & \multicolumn{2}{c}{10.81}                                                                             & 0.4393                                                                               & 3.0864                                                                       & 2.8424                                                                       &                   &                   \\
640$\;ms$                       & 15.83                              & \multicolumn{2}{c}{12.14}                                                                             & 0.4419                                                                               & 3.1041                                                                       & 2.8808                                                                       &                   &                   \\
960$\;ms$                       & 16.34                              & \multicolumn{2}{c}{12.79}                                                                              & 0.4435                                                                               & 3.1141                                                                       & 2.8993                                                                       &                   &                   \\
1280$\;ms$                      & 16.57                              & \multicolumn{2}{c}{13.35}                                                                              & 0.4436                                                                               & 3.1166                                                                       & 2.9045                                                                       &                   &                   \\
1600$\;ms$                      & 16.75                              & \multicolumn{2}{c}{13.49}                                                                             & 0.4441                                                                               & 3.1203                                                                       & 2.9109                                                                       &                   &                   \\
1920$\;ms$                      & 16.85                              & \multicolumn{2}{c}{13.87}                                                                             & 0.4445                                                                               & 3.1232                                                                       & 2.9148                                                                       &                   &                   \\
2240$\;ms$                      & 17.02                              & \multicolumn{2}{c}{14.34}                                                                             & 0.4448                                                                               & 3.1234                                                                       & 2.9200                                                                       &                   &                   \\
2560$\;ms$                      & 17.17                              & \multicolumn{2}{c}{14.84}                                                                             & 0.4447                                                                               & 3.1234                                                                       & 2.9217                                                                       &                   &                   \\
2880$\;ms$                      & 17.23                              & \multicolumn{2}{c}{15.31}                                                                             & 0.4450                                                                               & 3.1267                                                                       & 2.9254                                                                       &                   &                   \\
3200$\;ms$                      & 17.16                              & \multicolumn{2}{c}{15.62}                                                                             & 0.4450                                                                               & 3.1253                                                                       & 2.9248                                                                       &                   &                   \\
4800$\;ms$                      & 17.52                              & \multicolumn{2}{c}{17.09}                                                                             & 0.4459                                                                               & 3.1311                                                                       & 2.9361                                                                       &                   &                   \\
10000$\;ms$                    & 18.05                              & \multicolumn{2}{c}{18.05}                                                                             & 0.4469                                                                               & 3.1394                                                                       & 2.9507                                                                       &                   &                  \\\bottomrule        
\end{tabular}
\caption{Numerical results of StreamSpeech on CVSS-C De$\rightarrow$En.}
\label{tab:numerical_deen}
\end{table*}

\end{document}